\documentclass[journal]{IEEEtran}
\pdfoutput=1
\usepackage{xtab}
\usepackage{multirow}
\usepackage{graphicx}
\usepackage{xcolor}
\usepackage{listings}
\usepackage{amsmath}
\usepackage{tabularx}
\usepackage{booktabs}
\usepackage{float}
\usepackage{balance}
\usepackage{stfloats}
\usepackage{longtable}
\usepackage{multicol}
\usepackage{hyperref}
\usepackage{cleveref}
\usepackage{xcolor} 
\ifCLASSINFOpdf
\else
\fi
\begin{document}
%
\title{The Evolution and Future Perspectives of Artificial Intelligence Generated Content}
%
%
%

\author{Chengzhang~Zhu,~\IEEEmembership{Member,~IEEE,}
        Luobin~Cui,~\IEEEmembership{Member,~IEEE,}
        ~Ying~Tang,~\IEEEmembership{Senior Member,~IEEE,}
        ~Jiacun~Wang,~\IEEEmembership{Senior Member,~IEEE}
\thanks{This work was supported in part by the Rowan University-Rutgers-Camden Board of Governors and the National Science Foundation under grant number 2121277. (\textit{Corresponding author: Ying Tang})}
\thanks{C. Zhu and L. Cui are with the Department of Electrical and Computer Engineering, Rowan University, Glassboro, NJ, 08028 USA (e-mail: zhuche95@students.rowan.edu, cuiluo77@students.rowan.edu).}
\thanks{Ying Tang is with the Department of Electrical and Computer Engineering, Rowan University, Glassboro, NJ, 08028 USA (e-mail: tang@rowan.edu).}
\thanks{Jiacun Wang is with the Department of Computer Science and Software Engineering, Monmouth University, West Long Branch, NJ, 07764 USA (e-mail: jwang@monmouth.edu).}

}
\maketitle

\begin{abstract}
Artificial intelligence generated content (AIGC), a rapidly advancing technology, is transforming content creation across domains, such as text, images, audio, and video. Its growing potential has attracted more and more researchers and investors to explore and expand its possibilities. This review traces AIGC's evolution through four developmental milestones, ranging from early rule-based systems to modern transfer learning models, within a unified framework that highlights how each milestone contributes uniquely to content generation. In particular, the paper employs a common example across all milestones to illustrate the capabilities and limitations of methods within each phase, providing a consistent evaluation of AIGC methodologies and their development. Furthermore, this paper addresses critical challenges associated with AIGC and proposes actionable strategies to mitigate them. This study aims to guide researchers and practitioners in selecting and optimizing AIGC models to enhance the quality and efficiency of content creation across diverse domains.
\end{abstract}


%
\IEEEpeerreviewmaketitle

\section{Introduction}
%
%
%
%
\IEEEPARstart{T}{he} advancement of Artificial Intelligence (AI) has transformed the way content is produced, leading to a promising technology called Artificial Intelligence Generated Content (AIGC). By harnessing AI algorithms, this technology leverages human creativity to complement the traditional content generation process, ultimately improving the quality of human-centered production work with applications across diverse digital mediums, including text \cite{openai2024gpt4}\cite{touvron2023llama}, images \cite{image1}\cite{gan}, audio \cite{music1}, and videos \cite{video1}. AIGC stands as a crucial development in information and data technology, redefining the efficiency and quality of creative outputs. 

As shown in Figure 1, many models and algorithms have been developed along the progressive trajectory of AIGC, with each technological advancement built upon the previous ones. Fresh endeavors persist in unfolding, leading to great tools \cite{openai2024gpt4} and systems \cite{rombach2022highresolution}\textcolor{red}{\cite{zhang2023pseai}} that support sustainable and adaptable content production across various sectors.   Part of the reason for so much interest in this development is the complexity of the problem and its many associated challenges. While data has been a fundamental drive to advance digital transformation across industries, its collection is often insufficient \cite{9994611}\cite{su15076032} or sometimes impossible 
\cite{paschek2022industry}. Data annotation is rather time-consuming and costly, leading to the available public datasets often being on a small scale. Examples include breast pathological images \cite{liu2022deep},  the MNIST dataset \cite{lecun1998mnist} for handwritten digits, and the CoNLL-2003 dataset \cite{tjong2003introduction} for named entity recognition. When significant manpower is added, it becomes harder to maintain consistency in tone, style, and quality. These inherent problems highlight the need for rapid, scalable, and consistent content production methods.


\begin{figure*}[ht]
\centering
\includegraphics[width=\textwidth]{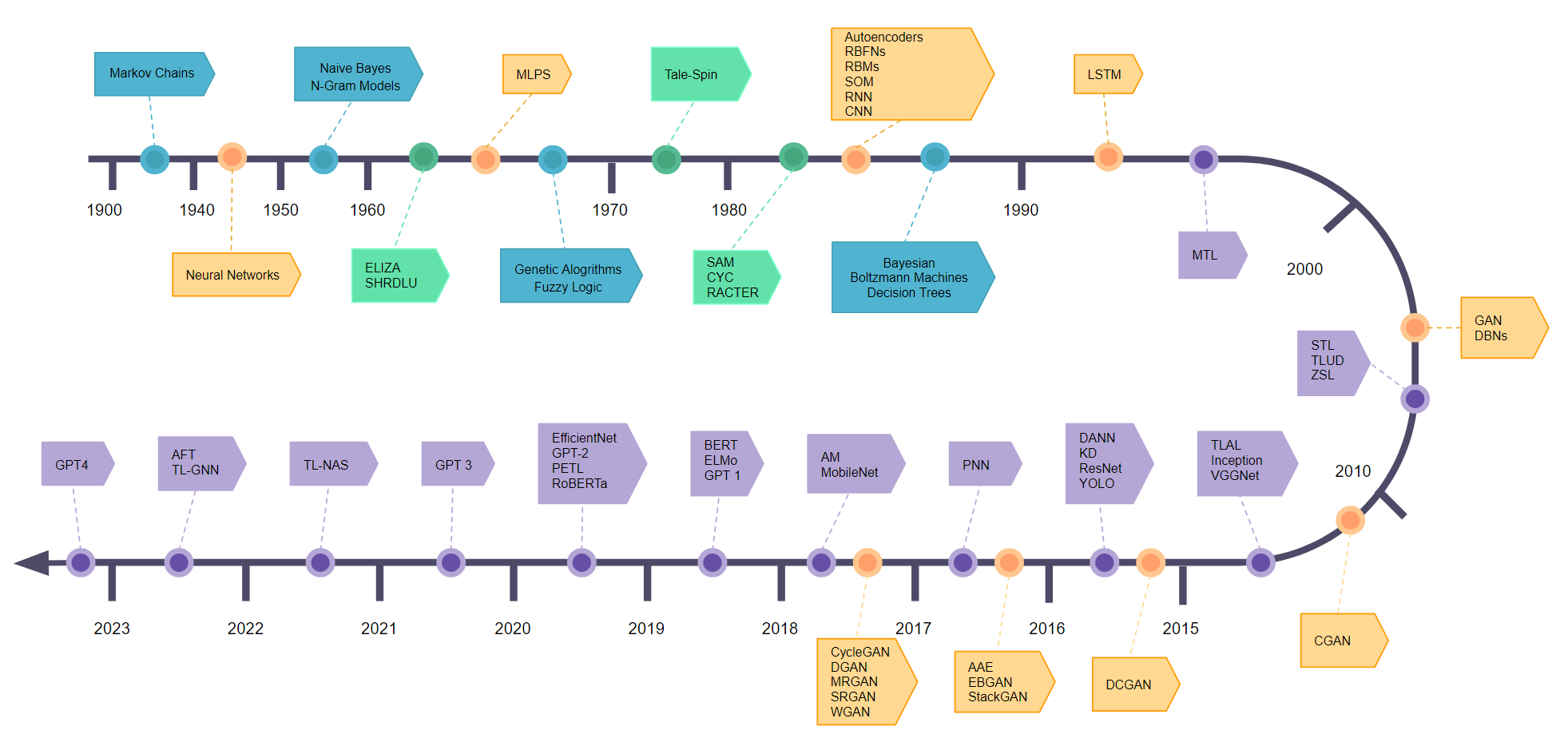}
\caption{Timeline of AIGC techniques.}
\label{fig:fig1}
\end{figure*}

Content generation approaches vary in terms of methods, and more importantly, in terms of limitations. Some have experimented with simple text generation to simulate human-like language interactions via rule-based methods \cite{ELIZA}\cite{SHRDLU}, while others capture statistical dependency between text elements in a sequence to predict the next element \cite{ngram1948}. These earlier attempts rely heavily on either experts’ rules established prior or a set of probabilistic measures, scoping their capabilities within predefined scenarios.  With the involvement of deep learning, the performance of generative models has been substantially improved, unlocking opportunities for a vast range of applications. Meanwhile, the critical need for trustworthiness in these models is also raised \cite{wu2023aigenerated}.

The evolution of AIGC progresses through multiple stages, beginning with the generation of foundational data, moving to data categorization and classification, and ultimately striving for human-like understanding. Although the processes involved in AIGC vary with the change in the intended outcomes, certain core steps form a robust framework across all applications. This paper conducts a comprehensive literature review to showcase the historical progression of AIGC within such a developmental framework. All AIGC methods are classified into four distinct categories - early rule-based systems, statistical models, deep learning models, and transfer \& pre-trained models - each adapts the foundational framework to its unique objectives and strengths. We intend to help researchers and practitioners make informed decisions when selecting AIGC models.

The contributions of this paper are twofold. First, this paper offers an in-depth review of the historical development of AIGC, mapping its journey from early rule-based systems to advanced neural network methodologies. More importantly, this review addresses both technological advancements and challenges, such as scalability and ethical dilemmas.

Second, unlike most of existing review papers that focus on either a specific application domain, such as text-to-3D generation\cite{ZhangGPTreview}\cite{text23D}, or a particular type of methodology such as Video GANs\cite{GANvideo}\cite{Cao2023AIGC}, our paper resembles \cite{cao2023} in its exploration of the historical evolution of AIGC but with even broader coverage. More importantly, this paper integrates a wider range of methodologies and applications into a cohesive framework that spans multiple phases of AIGC development. A unique contribution of this work is the use of a consistent example throughout, which not only enables readers to follow the evolution of AIGC methods but also makes it easier to understand and compare the differences among various approaches within a practical context. While we share the same view of the challenges facing AIGC, such as bias, ethical concerns, and scalability of models \cite{Wang_2023_review}, this paper extends the discussions by proposing research directions to tackle these issues. With these directions outlined, this paper contributes to the ongoing discourse on responsible AIGC development in real-world applications.

The rest of the paper is organized as follows. Section II provides the criteria for literature inclusion in this review.  A detailed exploration of AIGC is given in Section III, presenting its core characteristics and advantages such as automation, consistency, and efficiency. Section IV delves into the early rule-based systems that lay the foundation for AIGC, followed by a discussion of the evolution towards statistical methods, deep learning models, and the integration of transfer and pre-trained models in Sections V, VI, and VII, respectively. Each section examines technological advancements and applications of these methodologies in real-world scenarios. Section VIII addresses the potential challenges and limitations of current AIGC technologies. Finally, Section IX concludes the review by summarizing the comprehensive insights gathered throughout the paper and suggesting future research directions.

\section{Criteria for Inclusion in the Literature Review}

\textcolor{red}{
To conduct a structured review of AIGC research, a clear literature search strategy is developed in alignment with our research objectives. The search begins with the identification of main keywords, including “Artificial Intelligence Generated Content,” “Generative AI,” and “Machine-generated Content". Additional terms such as “transformer-based content generation” and “neural network content creation” are also incorporated to expand the scope of the relevant materials.
}

\textcolor{red}{
Six major academic databases are then selected for the search: IEEE Xplore, ACM Digital Library, ScienceDirect, SpringerLink, arXiv, and Google Scholar. To ensure comprehensive coverage of the AIGC's historical development, additional sources are also considered, including MDPI, Frontiers, Wiley, Taylor and Francis. 
}

\textcolor{red}{
The initial query returns thousands of articles. To manage this volume, a two-step screening process is applied:
}

\begin{enumerate}
    \item \textcolor{red}{\textbf{Preliminary Screening.} Two researchers review the title and abstract of each article independently. Papers are excluded if they are not peer-reviewed, not written in English, duplicated, or not directly relevant to AIGC—e.g., works that only mention generative tools without detailed elaboration of underlying models, algorithms, or core applications.}
    
    \item \textcolor{red}{\textbf{Full-Text Evaluation.} After the initial screening, a total of 274 papers remain and then undergo full-text evaluation. Backward and forward citation searches are also performed to reduce the risk of omitting important studies. As part of the step, all papers are categorized into four developmental stages of AIGC evolution as mentioned in the Introduction: early rule-based systems, statistical models, deep learning models, and transfer learning and pre-trained models. Within each category, papers are reviewed chronologically to capture key milestones in the field’s progression and then selected based on the following two criteria: 
    \begin{itemize}
   \item \textit {Focused scope:} The work explicitly develops or analyzes AIGC models or algorithms;
   \item \textit{Technical depth:} The work presents a clear and detailed explanation of model design, training procedures, and evaluation experiments.
\end{itemize}
}
\end{enumerate}

\begin{figure*}[ht]
\centering
\includegraphics[width=0.6\textwidth]{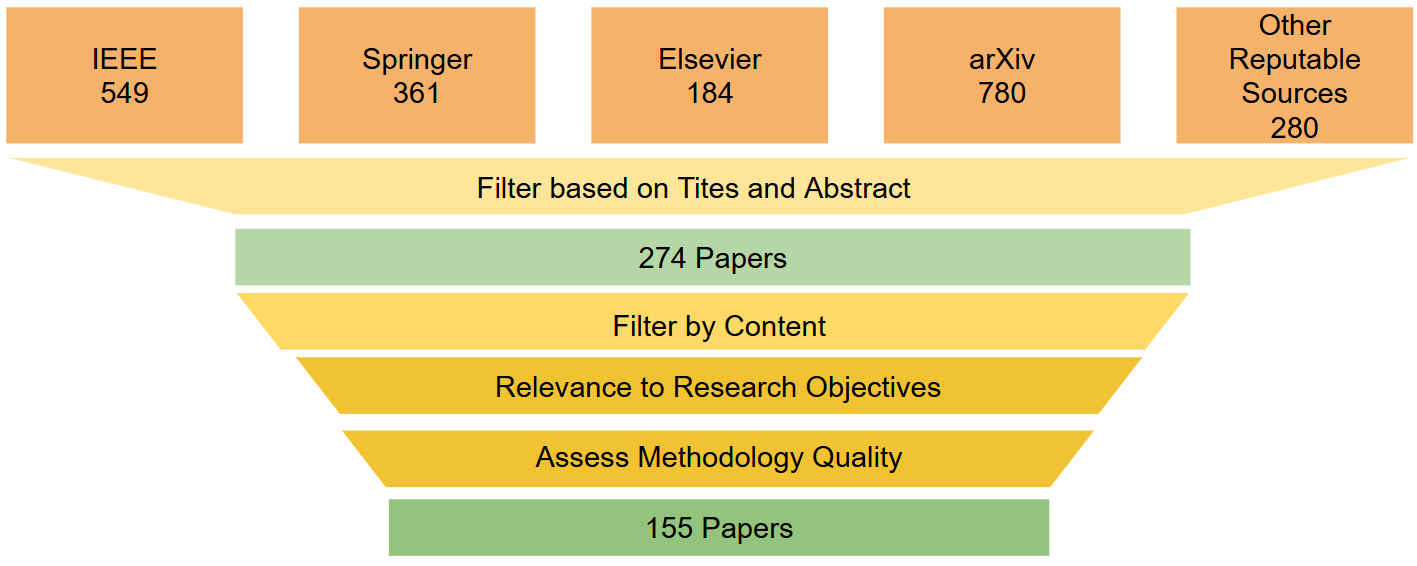}
\caption{Literature search and screening process}
\label{fig:fig.1}
\end{figure*}

\textcolor{red}{
For each milestone topic, if multiple articles are identified on the same subject, priority is always given to papers with higher citation counts or those published in journals with higher impact factors. When these papers are from the same research group, only the most foundational or representative work is retained to avoid redundancy. However, if they originate from different research groups, multiple works might be included, provided that each features a distinct technical perspective. The full screening process and the number of papers retained at each stage are presented in Fig.~\ref{fig:fig.1}.
}

\textcolor{red}{
Based on the aforementioned literature search strategies, our review spans the past century, primarily encompassing all relevant publications released prior to 2024. Although the formal search window ends in 2023, a few key articles released in early 2024 are also included—particularly those that discuss emerging challenges and future directions in AIGC development. In total, 155 high-quality papers are selected for detailed analysis in this review. For each of them, key information is extracted, including the generative task type, model architecture, datasets, reported limitations, and suggestions for future work.}

\begin{figure*}[ht]
\centering
\includegraphics[width=0.6\textwidth]{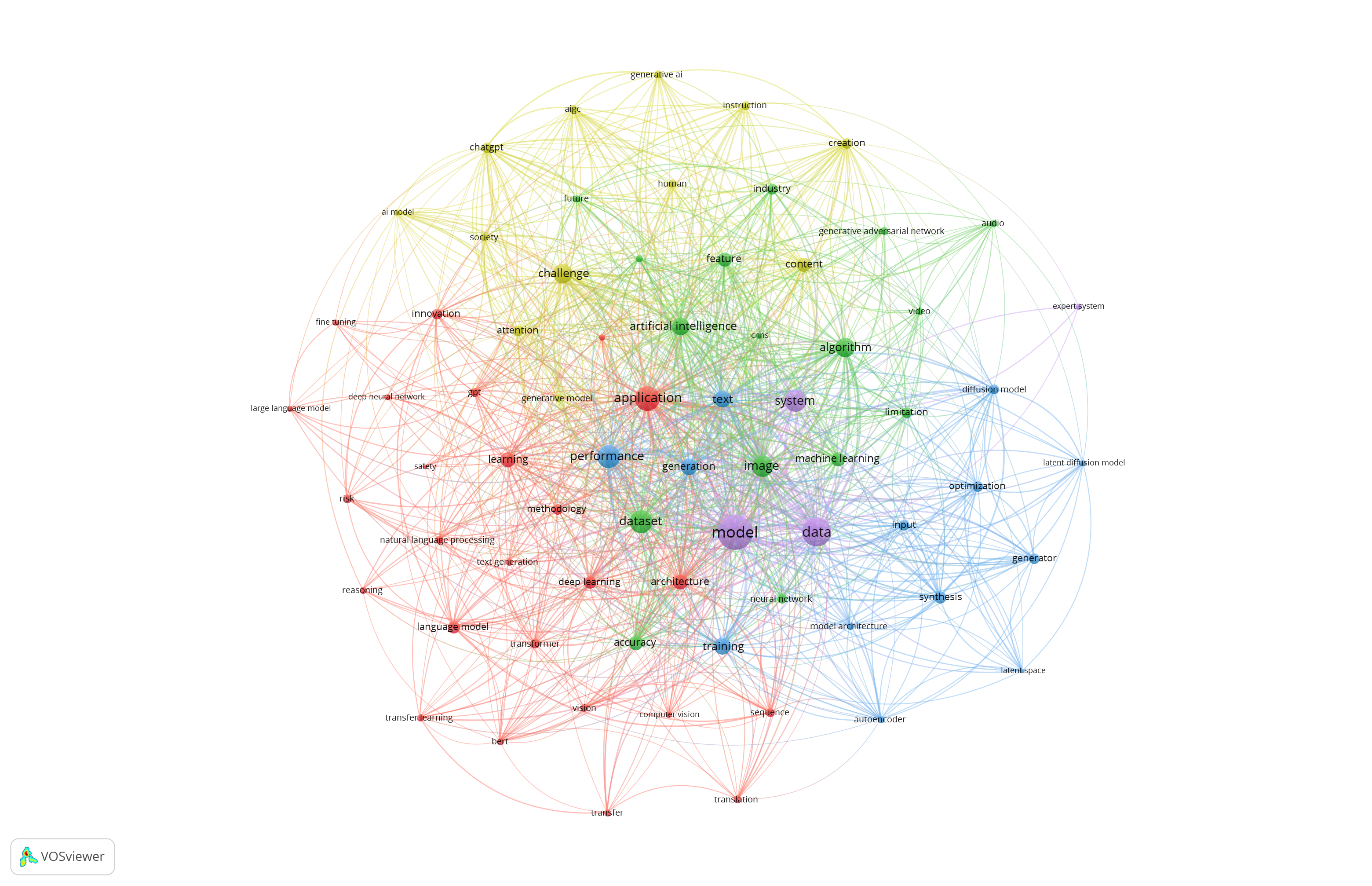}
\caption{Keyword co-occurrence networks after screening.}
\label{fig:fig.VOS}
\end{figure*}

\textcolor{red}{To validate our search process, VOSviewer is utilized to visualize keyword co-occurrence analysis, as shown in Fig.~\ref {fig:fig.VOS}. It is clear that the key terms are well-clustered into several coherent groups, highlighting the comprehensiveness and relevance of our retrieved literature. For instance, the red cluster focuses on the foundational models and methodological underpinnings of AIGC, with keywords like "deep neural network," "Transformer," "BERT," "deep learning," " transfer learning," and "fine-tuning." The blue cluster emphasizes generative approaches, such as "autoencoder," "latent space," "synthesis", and "diffusion model." Meanwhile, the green cluster centers around general AI and content generation applications, including keywords like "algorithm," "feature", "image," "video", and "audio". Overall, the structured distribution of terms with the presence of both foundational and applied keywords confirm that our search strategies successfully captured a diverse and balanced spectrum of relevant literature.}

\begin{figure*}[ht]
\centering
\includegraphics[width=0.6\textwidth]{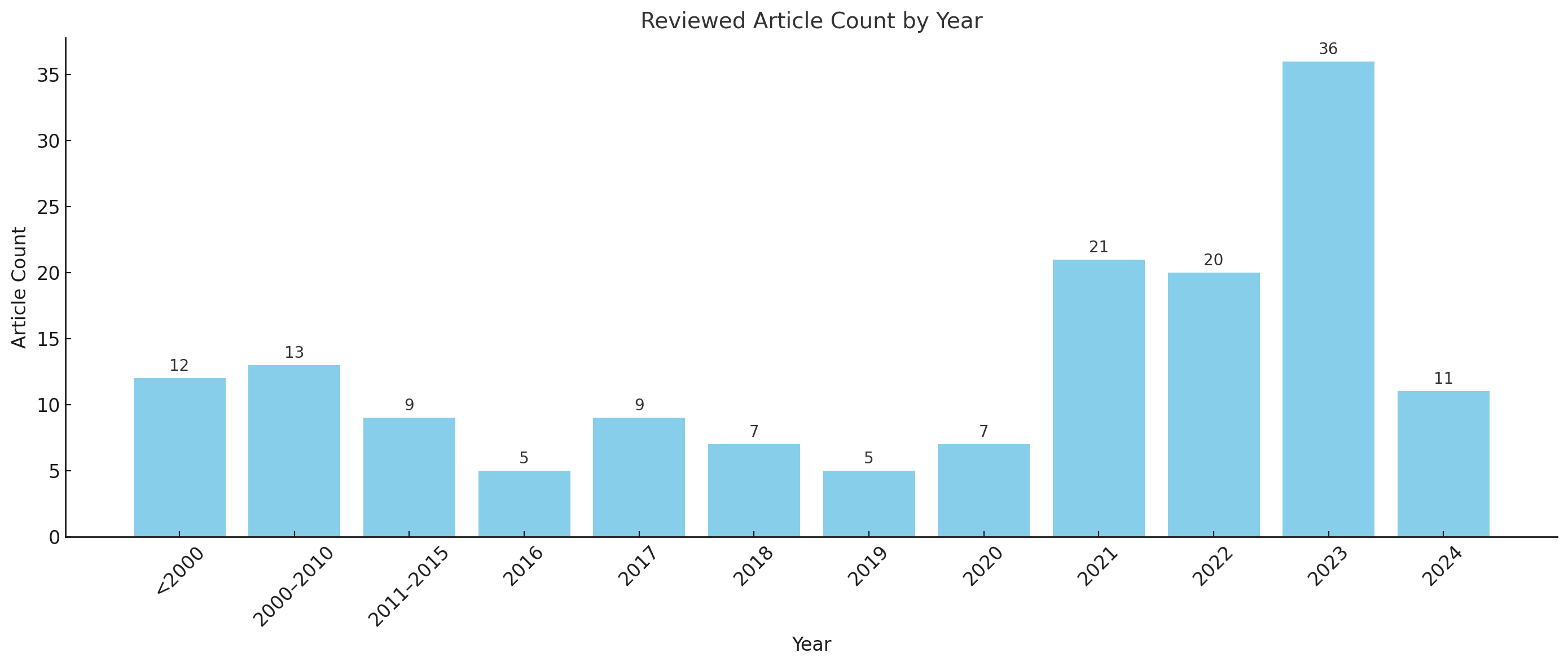}
\caption{Reviewed article count by year.}
\label{fig:year}
\end{figure*}

\textcolor{red}{Finally, the distribution of all selected articles by publication year is presented in Fig.~\ref{fig:year},  illustrating the temporal trends in the development of AIGC-related research. As illustrated, the number of publications on AIGC has increased significantly in recent years, particularly after 2020. This trend coincides with the emergence and widespread adoption of transformer-based models and large-scale generative frameworks (e.g., GPT-4\cite{openai2024gpt4}). The sharp rise in 2022 and 2023 suggests a growing academic interest and a transition from foundational research to application-oriented studies across multiple modalities, including text, image, audio, and video generation.}

\section{General AIGC Technique Framework}

AIGC is an innovative way of autonomously generating diverse forms of content by algorithms, machine learning models, or other artificial intelligence techniques. The main characteristic of AIGC is its reliance on data input and pre-generated models to produce content, requiring less or no human intervention. Compared to manual creation, AIGC has distinguished advantages in terms of automation, scale, and efficiency \cite{KUMAR2021}. It also ensures consistency, customization capability \cite{tong2021janus}, and cost-effectiveness \cite{jafari1994rulebased}. These strengths have continuously motivated researchers \cite{liu2021tracking} to advance AIGC technologies for decades, leading to a variety of applications as shown in Table \ref{tab:aigc_applications}. Beginning with basic text generation \cite{text1}, AIGC has now expanded to include video, audio \cite{GMM_audio}, and 3D model \cite{sd-3d} generation for applications in fields such as education \cite{Chen2024}, healthcare \cite{chen2023educationai}, entertainment \cite{li2023medicalai}, and many others.

\begin{table}[ht]
\centering
\caption{Overview of AIGC Applications}

\label{tab:aigc_applications}
\resizebox{1\linewidth}{!}{
\begin{tabular}{|l|l|}
\hline
\textbf{Category} & \textbf{Branches} \\
\hline
\multirow{4}{*}{Textual Generation} & Article Writing\cite{9853785} \\
 & Chatbots\cite{9257001}\cite{10125121}\\
 & Language Translation\cite{10257015}\\
 & Text Summarization\cite{10115423}\\

\hline
\multirow{5}{*}{Visual Generation} & Image Creation\cite{DBLP:journals/corr/abs-2102-09109}\cite{10005618}\\
 & Video Production\cite{9540056}\cite{9028743} \\
 & Graphic Design\cite{engawi2022impact} \\
 & 3D Modeling\cite{xu2023dream3d}\cite{wang2023gan3d}\cite{zhao2023text3d}\\
 & Digital Art Generation\cite{anantrasirichai2022artificial}\cite{10124952} \\
\hline
\multirow{3}{*}{Audio Generation} & Music Composition\cite{tan2021survey} \\
 & Voice Synthesis\cite{zhang2022visinger} \\
 & Audio Editing\cite{wang2023audit} \\
\hline
\multirow{3}{*}{Code Generation} & Automated Scripting \cite{sun2021experience} \\
 & Algorithm Design\cite{amuru2023aiml} \\
\hline
\multirow{4}{*}{Multimodal Generation} 
 & VR/AR Content Creation\cite{wang2020vr} \\
 & Interactive Media\cite{anantrasirichai2022ai} \\
 & Simulation Generation\cite{9583606}\cite{9362238} \\
\hline
\multirow{3}{*}{Data Generation} & Data Synthetic\cite{kim2023synthetic}\cite{lingo2023exploringpotentialaigeneratedsynthetic} \\
 & Data Augmentation\cite{singh2023augmentation}\\
\hline
\multirow{4}{*}{Data Modeling and Analytics} & Predictive Analytics\cite{kumar2023industry4}  \\
 & Anomaly Detection\cite{garcia2023anomaly}\\
 & Operational Optimization\cite{lee2023aior}\\
 & Recommendation Systems\cite{Zhang2021}\\
\hline
\end{tabular}}
\end{table}

\begin{figure}[ht]
\centering
\includegraphics[width=0.5\textwidth]{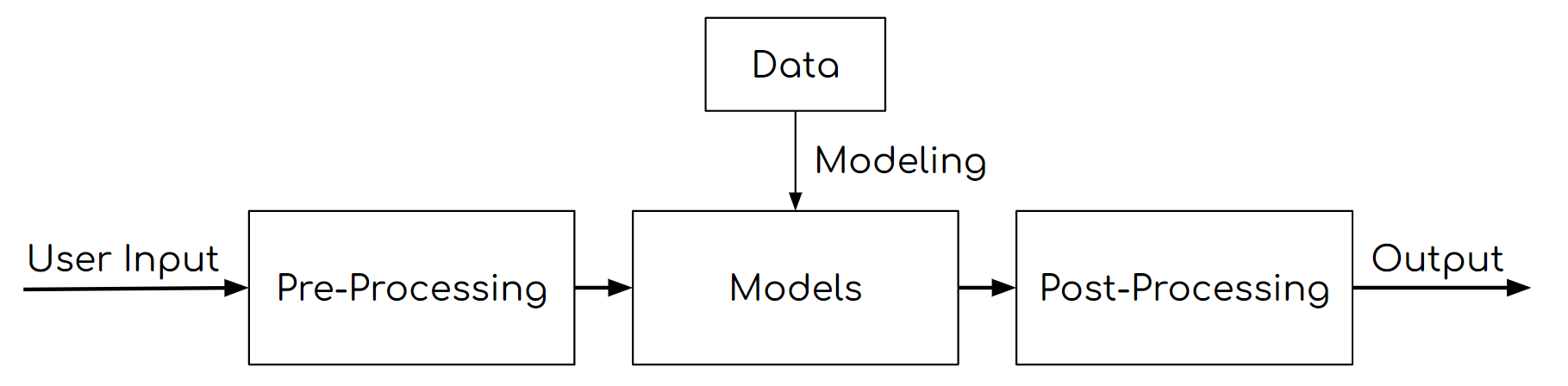}
\caption{General AIGC technique framework.}
\label{fig:fig2}
\end{figure}

Behind such a wide range of application domains, technologies underpinning AIGC are even more diverse. As mentioned earlier, this review divides the evolution of AIGC into four key milestones: early rule-based systems, statistical models, deep learning models, and transfer and pre-trained models. These milestones are intertwined and mutually reinforcing. Upon closer examination, it is evident that they share similar core steps, as shown in Figure \ref{fig:fig2}.

The flow of an AIGC system begins with user input that goes through a pre-processing phase to convert from its raw, naturally human-understandable form into a structured, machine-interpretable format \cite{fan2021review}. This usually involves techniques such as natural language processing (NLP) \cite{collobert2011natural} for text data or feature extraction for visual \cite{park2021image} or auditory data, with methods like decision trees and pattern matching bridging input rules to algorithms.

After preprocessing, the data is fed into models that are carefully selected, constructed, and trained to match the features of the input data and the desired output. These models, ranging from simple decision trees to complex neural networks, recognize patterns, make decisions, or generate new content based on the structured input data they receive.

The output generated by models is usually not in the final form required by the user. Therefore, a post-processing step is necessary. In this stage, the output of a model is refined and transformed into a form that is easy for the user to understand and meets the user's needs. This may involve converting the data back into a natural language \cite{zhang2023survey}, creating visual representations \cite{seeram2008image}, or simply formatting or enhancing the content to meet user expectations and ensure a high-quality experience.

These sequential steps, each serving a unique role, contribute collaboratively to the robustness and efficacy of the content generation process. A deep understanding of these steps in each developmental milestone of AIGC helps us gain insights into where AIGC has been and where it is heading. To that end, this paper uses the same example of “generate a research question with the keywords: artificial intelligence, healthcare, and ethical implications” across all milestones. This unified reference point not only demonstrates the unique advantages of the methodologies at each milestone, such as the transparency of rule-based systems, the data-driven nature of statistical methods, the automation capabilities of deep learning, and the flexibility of transfer learning, but also makes comparisons between approaches more straightforward and clear. Ultimately, the paper aims to assist researchers in better grasping each method's operational process, enabling them to choose a proper one for their practical applications or combine different ones for optimal results. 

\textcolor{red}{The choice of text generation as the recurring example stems from natural language generation’s earlier development and structured theoretical lineage. Considering the conceptual overlaps among generative models and frameworks across various modalities, especially in how they manage data representations and content synthesis, highlighting text generation serves to leverage a mature example for clarity, without narrowing the scope of our review to text alone.}

\section{Early Rule-based Systems}
\begin{figure}[ht]
\centering
\includegraphics[width=0.5\textwidth]{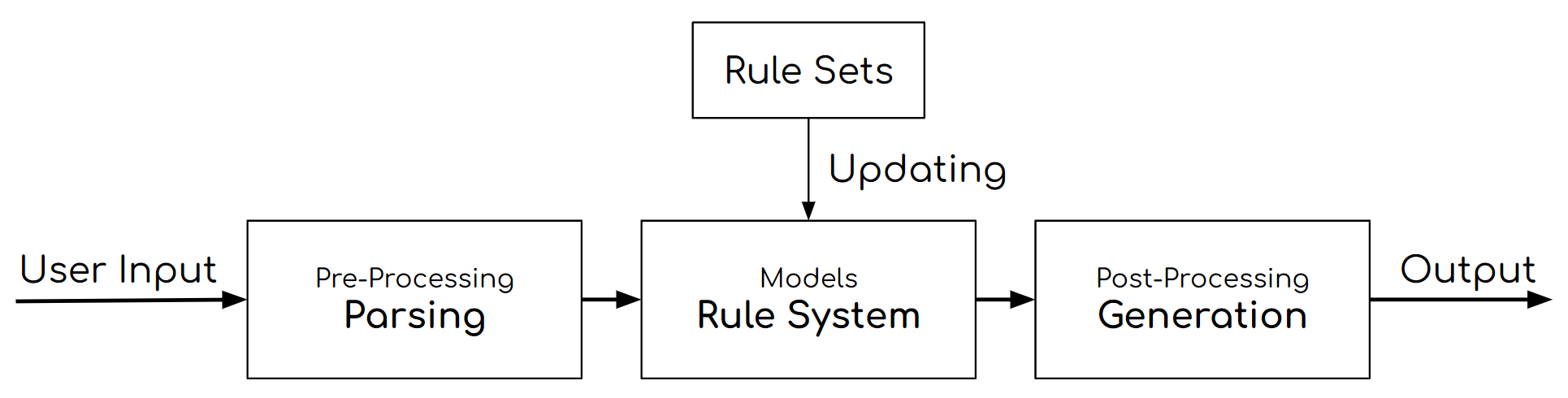}
\caption{Early rule-based system prototype.}
\label{fig:fig3}
\end{figure}

The roots of AIGC can be traced back to the 1950s \cite{juang1991hidden}\cite{marcille2022gaussian} when early rule-based systems lay the foundation for its development. These systems employ a set of expert rules defined a priori to guide content generation. Figure \ref{fig:fig3} presents the specific framework for rule-based systems within the context of AIGC discussed in the previous section.

\subsection{Fundamentals}

Early rule-based systems primarily focus on texts, using NLP techniques to enable computers to understand human inputs, in which the preprocessing step is attributed to ‘parsing’. The common methods include tokenization, stemming, stop word removal, rule-based parsing, and entity recognition. Tokenization breaks down the input data into small pieces, like words. Doing so converts a continuous stream of text into a manageable sequence of tokens, making it easier for algorithms to interpret. Stemming reduces words to their basic form, like 'running' to 'run'. This ensures consistency across similar words, reducing complexity with fewer unique words. In the step of stop word removal, common yet nonessential words like 'the' and 'is' are removed. This step helps the system focus on meaningful words only and reduces noise in the data. Rule-based parsing applies grammar rules to understand sentence structure, aiding in the interpretation of word relationships and the comprehension of text syntax and semantics. In addition, Entity recognition helps spot and categorize keywords, enabling the quick identification of important information such as the text's main subjects. These steps streamline the complexity of natural language, making it more structured and manageable for analysis and encoding. 

The user input, after pre-processing, is then screened through a set of rules pre-defined by domain experts, often referred to as expert rules \cite{Cowan2001} to guide grammar, style, and, more importantly, the logic of text generation. This screening process is termed an $inference$ that controls which set of expert rules are applied and how they are applied efficiently and accurately to user inputs. One such mechanism is the decision-making structure \cite{rowe1983top}\cite{hunt1992induction}, utilizing a flowchart-like or "if-then" reasoning to generate text responses. It searches linguistic, formatting, or narrative features or patterns in the input data that match some of "if" conditions to determine the proper response. For example, if the system detects the word "father" in user input, it applies rules associated with the "family" category to construct a response. Of course, there is always a default rule for situations where the system fails to recognize any matching patterns in user inputs, leading to a possible predefined response like "I can't understand".

The role of post-processing in the rule-based phase is pretty straightforward. It usually consists of some content enrichment processes, such as sentence re-organization, spelling, and grammar corrections, to improve the clarity and fluency of generated results. The decision to enhance the consistency or diversity, and depth of the language used is made through content formatting and lexical enrichment.


\subsection{Example}
Early rule-based systems find applications in text generation. For example, Eliza 
 is designed as a chatbot for psychotherapist conversations, SHRDLU  advances semantic analysis, and Tale-Spin \cite{TALESPIN} can create poetry and stories. Here, we detail how Eliza responds to the user input “generate a research question with the keywords: artificial intelligence, healthcare, and ethical implications" to showcase the operations of early rule-based systems.
 
\begin{figure}[ht]
\centering
\includegraphics[width=0.5\textwidth]{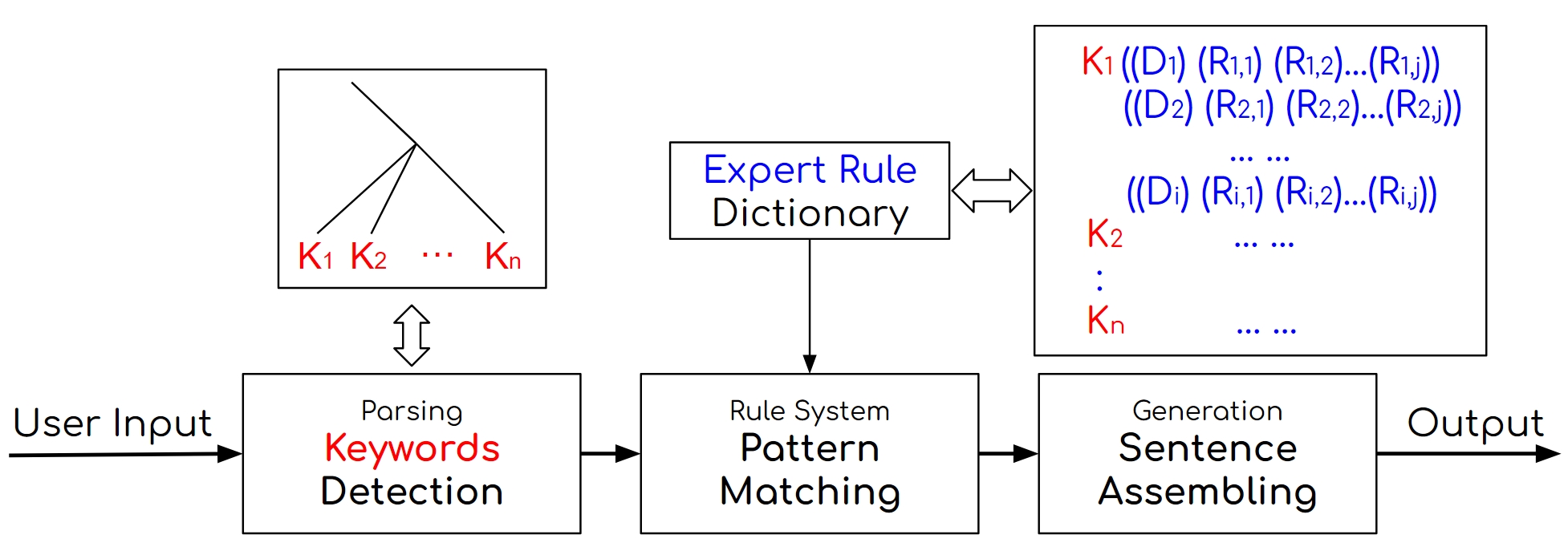}
\caption{Eliza system design.}
\label{fig:fig4}
\end{figure}

\begin{figure}[ht]
\centering
\includegraphics[width=0.5\textwidth]{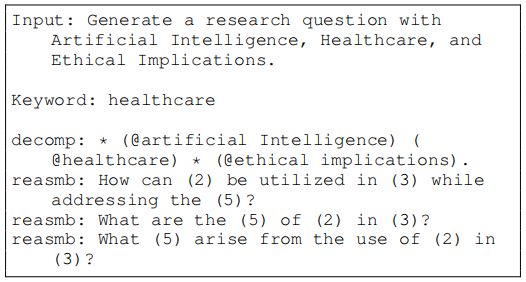}
\caption{An Eliza rule sample.}
\label{fig:eliza}
\end{figure}

 As shown in Figure \ref{fig:fig4}, Eliza first parses the user’s input to identify a number of keywords $K$. In our case, three keywords are found, which are “Artificial Intelligence”, “Healthcare”, and “Ethical Implications”. They are then matched with a dictionary of preset rules for keyword matching. Each rule consists of a ’keyword’, a corresponding set of decomposition patterns $D$, and a set of reassembly policies $R$ used to generate system responses. Note that the dictionary organizes the rules based on categorized keywords. For instance, “Artificial Intelligence”, “Healthcare”, and “Ethical Implications” fall under the categories of technology, background, and application, respectively. During decomposition, each keyword is tagged with a “@”, and the rest of the user input is replaced by “*”. Based on their order in the sentence, an index number is assigned to tagged keywords and other phrases. In our example, the original input is decomposed as * (@artificial Intelligence (2)) (@healthcare (3)) * (@ethical implications (4)), where the two * represent “Generate a research question with” and “and” with the indices of 1 and 5, respectively. As for the set of reassembly rules, they are ordered according to priority. To avoid generating the same response repetitively, a reassembly rule, once chosen, is moved to the end of the list for future interactions. With these policies in place, the output of Eliza for our case is “What are the ethical implications of Artificial Intelligence in Healthcare?” as shown in the following rule sample Figure \ref{fig:eliza}.

Early rule-based systems offer a high degree of transparency, as the defined rules ensure consistent outputs. These systems perform very well when expert knowledge is accurate and complete \cite{Rule_TSMC}. Additionally, system updates can be quickly done by tweaking the rules. However, as rule sets grow, scalability in terms of rule management and navigation becomes an issue. With rules being static, under-responsiveness is unavoidable, especially when encountering unfamiliar scenarios. All of these limitations call for the development of new approaches, leading to the rise of statistical methods.

\section{Statistical Methods}

Statistical methods analyze large datasets to identify patterns, trends, and probabilities, relying on data-derived "rules" rather than expert-defined ones in early rule-based systems. By using probabilistic models, statistical methods estimate the likelihood of various outcomes, which are well-suited to handle dynamic and ambiguous real-world data.

\begin{figure}[ht]
\centering
\includegraphics[width=0.5\textwidth]{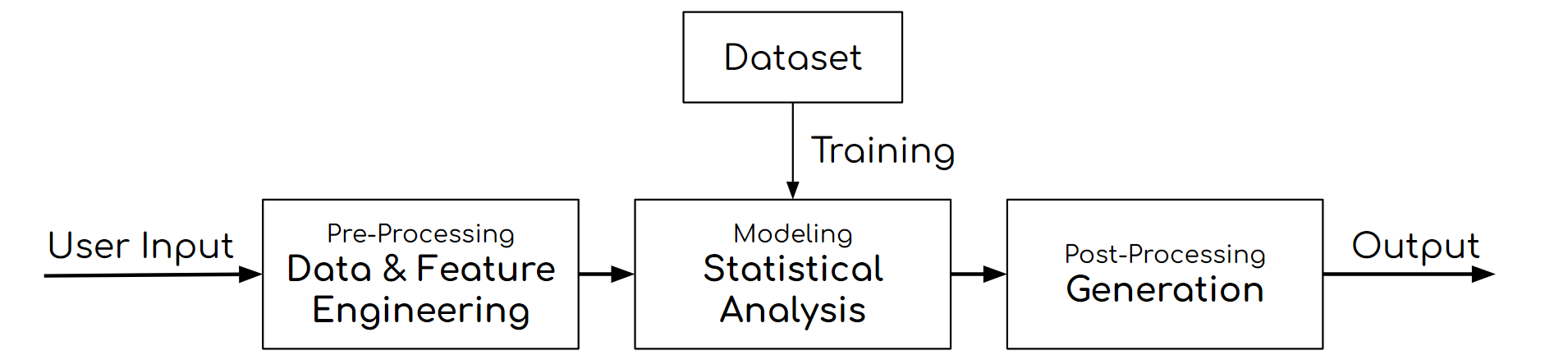}
\caption{Statistical methods prototype.}
\label{fig:fig5}
\end{figure}

\subsection{Fundamentals}

As shown in Figure \ref{fig:fig5}, preprocessing for NLP tasks in statistical methods resembles early rule-based systems, including steps like tokenization, lemmatization (stemming), and stop-word removal. However, what sets it apart is the calculation of term frequency and inverse document frequency to obtain word importance in the user’s input. This step is also known as feature extraction. Additionally, normalization is used to scale features to a uniform range to ensure equal contributions from all inputs to the model's adaptation.

The key aspect of statistical methods is their probabilistic models, which capture uncertainty and ambiguity through continuous learning. Initialized with human-set parameters, these models are trained on selected datasets and fine-tuned their parameters using algorithms such as regression, classification, or clustering. Unlike rule-based models that follow strict 'if-then' logic, statistical methods rely on trained models to predict outcomes by adapting to patterns and trends in data. Table \ref{tab:ml_methods} shows the five most representative statistical methods.

\begin{table}[ht]
\centering
\caption{Overview of Machine Learning Methods}
\label{tab:ml_methods}
\begin{tabular}{|l|l|l|}
\hline
\textbf{Method Name} & \textbf{Year} & \textbf{Authors} \\
\hline
N-Gram Models & 1948 & Claude Shannon\cite{ngram1948} \\
\hline
Hidden Markov Models (HMM) & 1972 & L.E. Baum et al.\cite{HMM1972} \\
\hline
Gaussian Mixture Models (GMM) & 1980 & L.P. Hansen\cite{GMM1982} \\
\hline
Restricted Boltzmann Machines (RBM) & 1986 & Paul Smolensky\cite{RBM1986} \\
\hline
Latent Dirichlet Allocation (LDA) & 2003 & D. M. Blei et al.\cite{lda2003} \\
\hline
\end{tabular}
\end{table}

In the realm of statistical language processing, N-gram models serve as a straightforward yet effective method for predicting the next item in a sequence based on the preceding N items. They are particularly valuable for tasks such as lexical selection and modifier ordering \cite{ngram_text} due to their ability to capture local context. Yet N-gram is effective for text generation and speech recognition \cite{ngram_audio}, it often faces several limitations, especially when dealing with long-range dependencies and language ambiguity, owing to its fixed context window. To address these limitations, researchers develop Hidden Markov Models (HMMs), which build upon N-grams by incorporating hidden states, allowing for more flexible sequence modeling. HMMs are crucial for handling temporal variations, making them indispensable in tasks like part-of-speech tagging and speech recognition \cite{HMM_Voice}\cite{6408207}. Nonetheless, HMMs are still constrained by their assumptions about state independence and often require extensive training data to achieve high accuracy.

As the need for more sophisticated sequence modeling grows, Gaussian Mixture Models (GMMs) are introduced to extend the capabilities of HMMs. By representing data with mixed distributions, GMMs excel in distinguishing complex patterns, such as those found in audio signal processing and text clustering. By effectively handling data variations \cite{GMM_text}, GMMs offer improved classification accuracy\cite{GMM_audio}. However, GMMs can be computationally intensive and may struggle with high-dimensional data without adequate preprocessing and feature selection.

In the pursuit of uncovering hidden patterns within data, researchers turn to Restricted Boltzmann Machines (RBMs). These models capture complex nonlinear relationships through their hidden units. Although typically categorized under machine learning due to their neural network architecture, RBMs are also considered statistical models because they employ probabilistic methods to model data distributions. RBMs are well-suited for tasks like topic modeling, summarization \cite{rbm_text}, and speech synthesis. Their generative nature allows them to uncover dependencies and structures in the data that simpler statistical models might miss \cite{rbm_audio}. Despite their potential, training RBMs can be challenging due to their sensitivity to hyperparameters and potential for slow convergence.

Building on the capabilities of RBMs, Latent Dirichlet Allocation (LDA) focuses specifically on topic modeling and document classification. LDA assumes that each document is a mixture of various topics, which are, in turn, mixtures of words. This hierarchical structure allows LDA to identify hidden topics within large text corpora \cite{LDA}, offering significant interpretability and utility in text analysis. However, LDA requires careful tuning and can be computationally demanding, particularly with large datasets.

\subsection{Example}

From the above discussion in Section V.A, it is evident that many statistical methods build upon N-grams, the simplest and most foundational technique. For instance, HMMs are an extension of N-grams with hidden states, adding a layer of abstraction to sequence modeling. Similarly, methods like GMMs, RBMs, and LDA each address specific limitations of N-grams and HMMs, offering more sophisticated approaches to handling ambiguity, complexity, and hierarchical structures within data. To illustrate the operational differences between the statistical methods and early rule-based models, we apply the same example that showcases the operations of early rule-based systems to N-grams - representative of statistical methods. The details are given below. 

In contrast to early rule-based systems that rely on predefined rules, N-grams generate “relationship rules” by learning from data. During training, common NLP steps are applied to the data, breaking each sentence into individual words. Special tokens, such as the start of sentences (SOS), end of sentences (EOS), and unknown words (UNK), are then added to the sentence. This step, transforming the original text into a sequence of discrete tokens, enables the model to analyze word relationships and predict word sequence.

\begin{figure}[ht]
\centering
\includegraphics[width=0.5\textwidth]{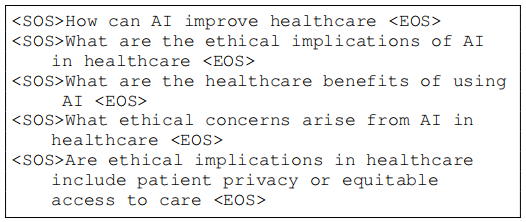}
\caption{N-Gram corpus}
\label{fig:ngram}
\end{figure}

The formation of n-grams depends on the choice of $n$ – the number of words in the sequence of interest. The model then calculates the frequency of each n-gram within the training data. For example, a unigram model calculates the probability of each word, while a 2-gram model calculates the probability of any two-word sequence appearing within the dataset. For this presentation, we use a bigram model as an example to illustrate how n-grams work. The 2-gram model is trained using a mini-corpus and then used to generate a response to the user. Figure \ref{fig:ngram} shows the mini corpus after tokenization, where bigrams can be extracted as pairs of consecutive words. Using the sentence “$<$SOS$>$ How can AI improve healthcare? $<$EOS$>$”, the bigrams are [($<$SOS$>$, How), (How, can), (can, AI), (AI, improve), (improve healthcare), (healthcare, $<$EOS$>$)]. Consequently, the likelihood of one word following another is computed as the total occurrences of the bigram divided by the frequency of the first word. For example, p(What $\mid$ $<$SOS$>$) =3/5. Using the trained bigram model, a response can then be generated. The decision on which word to start the response depends on how the model is set up and the context provided. The most straightforward way is the start token approach given that the start-of-sentence special token $<$SOS$>$ is included in each sentence of the training dataset. Typically, the system formulates a response from one of the most probable words following $<$SOS$>$. For our case, “What” frequently follows $<$SOS$>$, the system then chooses “What” as the first word. The same logic applies to generate a word after a word until $<$EOS$>$ appears, resulting in the response “What are the ethical implications of AI in healthcare”.

Traditional statistical models play a crucial role in the development of AIGC. However, since they rely on hand-crafted features or assumed data distributions, they work well on small, well-structured datasets. In other words, statistical methods struggle when dealing with large, high-dimensional, and unstructured data, where complexity and variability increase. This paves the way for deep learning to take a leading role in AIGC.

\section{Deep Learning Methods}

\begin{figure}[ht]
\centering
\includegraphics[width=0.5\textwidth]{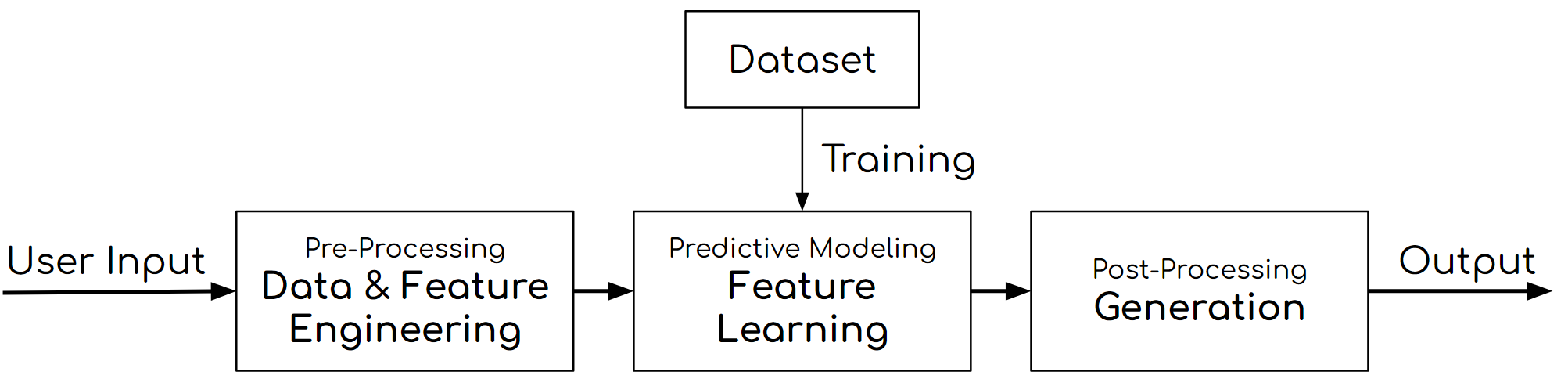}
\caption{Deep learning prototype.}
\label{fig:fig7}
\end{figure}

Neural networks and deep learning models mark a major leap forward in AIGC \cite{Sarker2021}, overcoming the limitations of statistical methods by better managing the growing volume, dimensionality, and complexity of modern data, especially unstructured data such as images, text, and speech. Mimicking the structure of the human brain, neural networks consist of layers of interconnected nodes (neurons). As data flows through the layers, the networks conduct a step-by-step inference process, automatically recognizing patterns and making predictions from input to output. Extended from this foundational structure, deep learning uses convolutional neural networks (CNNs), in which the nodes at each layer are organized into overlapping clusters so that each cluster feeds data to multiple nodes in subsequent layers. This design enhances the network's capability to process high-dimensional data to learn and capture complex relationships more effectively.
As shown in Figure \ref{fig:fig7}, deep-learning-based AIGC shares similarities with statistical methods in terms of feature engineering for user input. What sets them apart is the process of selecting, modifying, and creating features. Deep learning automates that process through training on diverse and large datasets, while statistical methods require human intervention. Notably, the quality and diversity of datasets are crucial for training deep learning models, as the data needs to represent a wide range of scenarios within the problem's scope. Thus, data augmentation is often applied to enhance both the size and variability of the training dataset, improving the model’s generalization ability. In deep learning, the backbone architectures include Recurrent Neural Networks (RNNs), CNNs, and Transformer \cite{Alzubaidi2021} as detailed in the following subsections.

\subsection{Convolutional Neural Networks (CNNs)}

CNNs are algorithms inspired by human vision and specialized in image processing. Similar to rule-based and statistical prototypes, they begin with raw signal ingestion, followed by preprocessing (e.g., initial determination of edges and orientation), figuration (e.g., determination of the object's shape, such as round), and finally object identification (e.g., recognizing it as a basketball). A CNN is composed of convolutional layers, pooling layers, and fully connected layers, which effectively addresses the challenges of large datasets and feature extraction in AIGC. Pooling layers, through downsampling, significantly reduce data dimensions and computational load, making feature detection robust to changes in scale and orientation. This design also helps mitigate the overfitting issues common in traditional statistical methods. By combining all learned features, the fully connected layers predict the final output, enhancing the quality and accuracy of generated content.

CNNs are built on the principles of local receptive fields, shared weights, and spatial or temporal invariance, which have helped propel AIGC into an era of rapid growth. Unlike earlier rule-based and statistical models, which often rely on handcrafted features and struggle with data in input, CNNs excel in automatically learning and extracting complex patterns. The use of local receptive fields allows CNNs to focus on specific regions of an input, capturing detailed features like edges and textures. Shared weights enable these networks to efficiently apply learned filters across the entire input, improving generalization and scalability. Spatial or temporal invariance, achieved through operations like pooling, ensures that CNNs can recognize patterns regardless of their position, addressing the limitations of previous models in handling diverse and complex inputs. This innovative structure makes CNNs particularly effective for content generation, as they provide robust, flexible, and scalable solutions for feature extraction and pattern recognition.

\subsection{Recurrent Neural Networks (RNNs)}

Just as its name indicates, the core idea of RNN is to utilize the cyclic connections of the network to capture temporal dependencies in sequence data. Compared to traditional neural networks, RNN introduces loops into the network, allowing it to take into account previous information when processing sequence data. This design makes RNNs well-suited for processing sequential data such as text, speech, and time series in AIGC tasks. The emergence of RNNs has enabled subsequent natural language processing tasks to learn the intrinsic inter-dependencies and patterns of sequential data well and make accurate predictions or decisions.

However, traditional RNNs suffer from the problems of gradient vanishing and gradient explosion, which make it difficult for the network to learn long-term dependencies. To address these problems, researchers have proposed some improved RNN structures such as Long Short-Term Memory Networks (LSTM) and Gated Recurrent Units (GRU) \cite{chung2014empirical}. These improved structures enable the network to better capture long-term dependencies by introducing a gating mechanism, which leads to more favorable results in many sequence processing tasks.

Overall, RNNs provide a solution for the processing of sequence data through their unique recurrent structure and state-keeping mechanism, making them an indispensable tool in many sequence processing tasks. 

\subsection{Transformer}
Despite the progress made in LSTM and GRU, issues with training speed and long-term dependency on RNN-related methods persist until the transformer architecture introduced by Google in 2017 \cite{attention}. By placing the attention mechanism at its core rather than as a supportive component, the model revolutionizes the parallel sequence processing of data. Typically, the mechanism weighs the importance of different elements in a sequence, allowing the model to efficiently capture long-range dependencies and focus on the most relevant parts.

\begin{figure}[ht]
\centering
\includegraphics[width=0.5\textwidth]{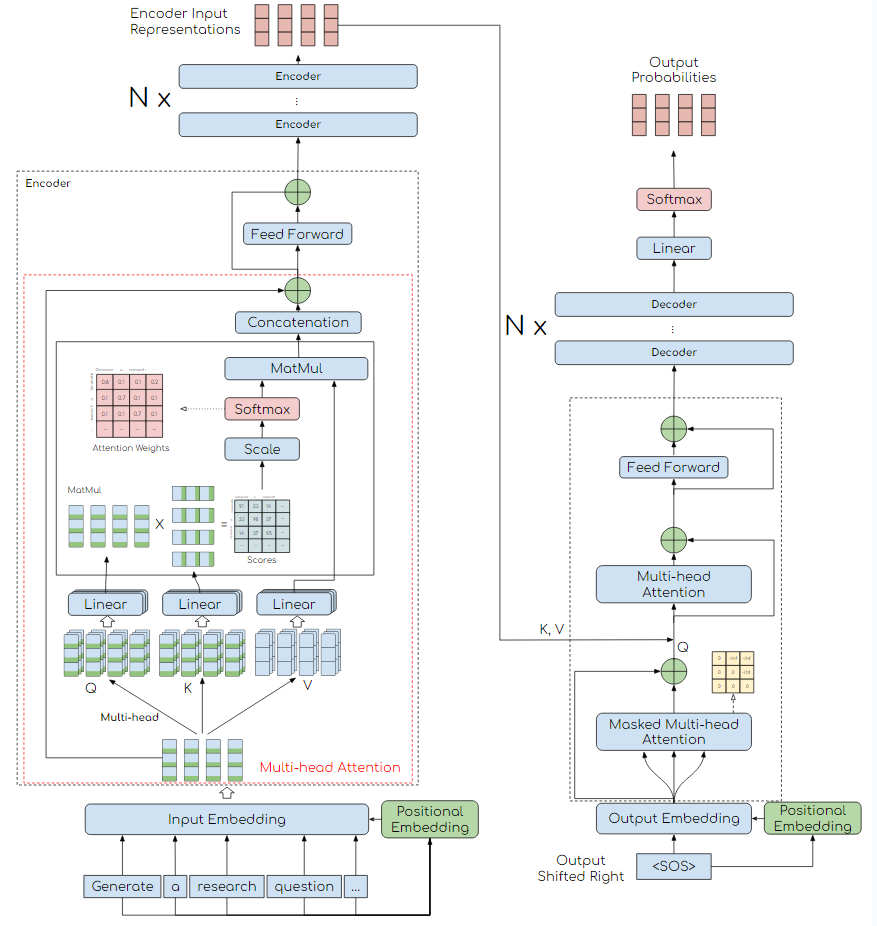}
\caption{Transformer structure.}
\label{fig:transam}
\end{figure}

As shown in Figure \ref{fig:transam}, the transformer consists of two main parts: multiple layers $j$ (i.e., \( j = N \)) of an encoder and a decoder that have very similar structures but serve different purposes. The former is responsible for encoding the input sequence into a context-rich representation, while the latter is used to generate new information in a sequence based on the given input and its encoded representation.

Before the input sequence is fed into the encoder, it is first tokenized, similar to that in early rule-based and statistical methods. What sets the transformer apart is the embedding process, mapping each token into a dense vector representation with a dimension of \( d_{\text{model}}\). Note that such dense vectors are generated from an embedding matrix, either derived from a pre-trained model (e.g., BERT \cite{Bert}, GPT) or trained from scratch during the model’s training stage. As stated earlier, the transformer model processes the entire input in parallel, and it does not have an inherent understanding of the sequence order of the tokens. Therefore, positional encoding is essential, as it is ultimately added to the input embedding to help the model gain the necessary sequence order information. While sine and cosine are the most traditional and widely used forms of positional encoding, many other approaches are available, such as learned embeddings and rotary positional embeddings\cite{roformer}, and their choice highly depends on the model architecture and the task requirements.

With the input embedding and positional encoding, the input sequence is transformed into an n-by-\( d_{\text{model}}\) matrix X. During the encoding phase, X is then split and processed through $m$ attention heads (i.e., $i=1,2,...m$) in parallel, each operating in a subspace (i.e., $d_{\text{head}} = \frac{d_{\text{model}}}{m}$) with its own set of Query (\( Q_i \)), Key (\( K_i \)), and Value (\( V_i \)) weight matrices of size \( d_{\text{model}}\)-by-\( d_{\text{head}}\) as shown in Eq. \ref{eq:qkv}. Thus, different aspects of the sequence can be attended simultaneously for richer representations through the attention calculation in Eq. \ref{eq:sofmax}. The outputs of all heads are concatenated and linearly transformed back to the original dimension (Eq. \ref{eq:attention}). Finally, the result passes through a feed-forward network (FFN), further refining its contextual understanding. The output of the FFN is then used as the input sequence to the next layer of the encoder. Note that the number of layers is an experiential parameter whose choice depends on model capacity, task complexity, and available computational resources. The more layers, the more complex patterns the model can capture, but the more resources are needed, the harder it becomes to train. It is recommended to start with a reasonable depth (e.g., 5 layers), and gradually adjust by increasing or decreasing based on validation performance. 

\begin{equation}
Q_i = XW_i^Q, \quad K_i = XW_i^K, \quad V_i = XW_i^V
\label{eq:qkv}
\end{equation}

\begin{equation}
Attention_i(Q_i,K_i,V_i)=softmax(\frac{Q_iK_i^{T}}{\sqrt{d_{head}}})V_i
\label{eq:sofmax}
\end{equation}

\begin{equation}
\begin{aligned}
& Attention(Q, K, V) = \text{Concat}(Attention_1(Q_1,K_1,V_1), \ldots, \\
& Attention_m(Q_m,K_m,V_m)) W
\label{eq:attention}
\end{aligned}
\end{equation}

The aforementioned attention mechanism applies to the multiple layers of the decoder in the transformer model as well with several key differences, given the decoder’s role in generating an output sequence based on the input. First, the output generation is usually done sequentially, predicting one token at a time. Thus, our interest here is the correlation between the next token and the previously generated ones. In other words, the self-attention in the decoder is masked to ensure that each token can only attend to earlier tokens in the sequence, but not the future ones. Second, the output generation must correspond to the given input sequence. Therefore, cross-attention is employed in the decoder to align its queries with the encoder’s output.

While initially introduced for NLP tasks, the modularity, and extensibility of the transformer model have made it a popular and powerful tool across a wide range of applications, such as modern large language models. Beyond excelling in text generation, it has recently been successfully extended to the area of computer vision, such as Vision Transformer \cite{vit}, showcasing its effectiveness in vision generation. 

\subsection{Variational Autoencoders (VAEs) and Generative Adversarial Networks (GANs)}

Built on these backbone techniques, the introduction of VAE and GAN represents further progress in addressing the limitations of earlier models, particularly in generating diverse and high-quality content. VAEs are effective at creating variations by learning data distributions. At the same time, GANs stand out for creating realistic images and videos through unique adversarial training, collectively pushing the boundaries of machine-generated creativity.

\subsubsection{VAEs}

The predecessor of VAEs is Autoencoder (AE) \cite{michelucci2022introduction}, an artificial neural network for unsupervised learning tasks. AEs are commonly used to handle tasks such as dimensionality reduction, feature learning, and data noise reduction and reconstruction. They compress input data into a low-dimensional latent space through an encoder and then reconstruct the data from the latent space using a decoder. 

The biggest difference between VAEs and AEs is the probabilistic layer VAEs introduce in the latent space, allowing them to sample and generate new data. Known for its generative power and distribution-driven coding, VAEs offer a rich and continuous latent space that is robust to input variations, making them good at handling image generation and style migration tasks. These advantages are attributed to the introduction of probabilistic encoding, stochastic latent space, and a two-part loss function that considers both reconstruction and regularization. However, VAEs sometimes struggle with fine detail, occasionally producing blurry outputs when dealing with highly complex data distributions.

\textcolor{red}{Building on the success of VAEs, recent applications such as Stable Diffusion \cite{sd} have emerged. These applications extensively utilize diffusion models, which generate high-quality images through an iterative denoising process. A generative diffusion model learns data distribution by gradually adding noise during training and reversing the process to denoise and generate samples. Early diffusion models, however, suffer from drawbacks such as high computational costs and slow generation speeds. By integrating the latent-space compression features of VAEs, Stable Diffusion overcomes these limitations, achieving higher-resolution, enhanced efficiency, greater diversity, and improved textual controllability in generated outputs.}

\subsubsection{GANs}

As an unsupervised learning, the emergence of GANs is no less than an industrial revolution in the evolutionary history of AIGC. Meeting the increasing demand for generated content, GANs improve the quality and diversity of generated content while also expanding labeled datasets, which is crucial for training other models.

GANs consist of two modules: the discriminator and the generator. The discriminator determines whether an input image is real (originating from the dataset) or machine-generated. The generator takes random noise and then uses an inverse convolutional network to create images. Both modules are improved iteratively against each other through adversarial losses so that the generator can produce more realistic data and the discriminator can more accurately distinguish between real and generated content.

Despite its potential, the original GAN framework faces challenges, such as training stability, limited controllability, high data requirements, and difficulties in achieving desired image quality. To address these limitations, researchers have proposed various extensions and refinements to the traditional GAN model. For instance, WGAN \cite{wgan} and LSGAN \cite{lsgan} improve training stability by replacing the binary cross-entropy loss function with Wasserstein distance and least squares loss, respectively. These modifications address issues like mode collapse in WGAN and gradient vanishing in LSGAN. To enhance controllability in producing outputs with specific attributes, researchers have explored approaches such as manipulating latent space as seen in InfoGAN \cite{infogan} and introducing conditional variables in CGAN \cite{cgan}. To reduce GAN's dependency on large labeled datasets, Pix2Pix \cite{pix2pix} uses paired datasets in supervised learning while CycleGAN \cite{cyclegan} and StarGAN \cite{stargan} employ unsupervised learning to avoid the need for paired data. In a similar fashion, researchers have sought to improve output resolution and diversity by scaling up the model capacity in BigGAN \cite{biggan} or introducing a control mechanism to manipulate image features for greater diversity in StyleGAN \cite{stylegan}.

\subsection{Example}
We choose the transformer as a prime example to showcase how deep learning methods work in response to the input of “generate a research question with the keywords: artificial intelligence, healthcare, and ethical implications".

\begin{figure}[ht]
\centering
\includegraphics[width=0.48\textwidth]{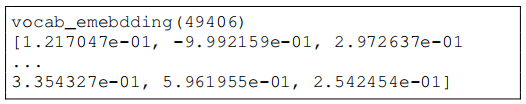}
\caption{$<$SOS$>$ vector.}
\label{fig:sos}
\end{figure}

\begin{figure}[ht]
\centering
\includegraphics[width=0.4\textwidth]{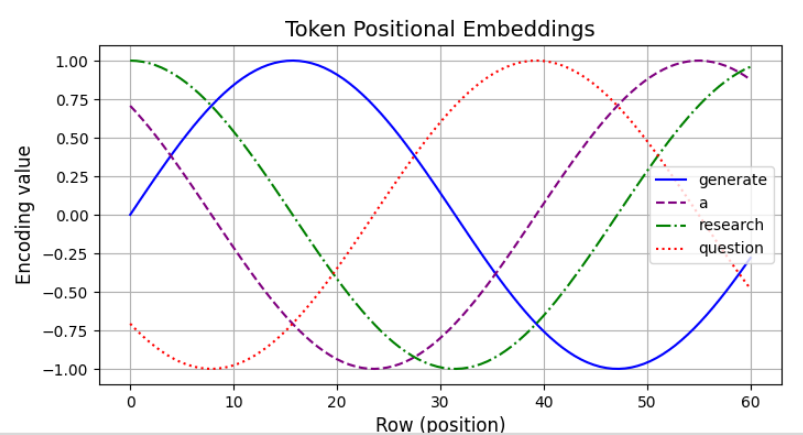}
\caption{Positional embeddings.}
\label{fig:pos}
\end{figure}

For this example, we train a corpus from scratch with 10,000 tokens and a dimensionality of 512. Using this trained corpus, the input sequence is then tokenized into 14 tokens, with each mapped to a 512-dimensional vector, resulting in an embedding matrix of size 16-by-512 with $<$SOS$>$, $<$EOS$>$ tokens. Figure \ref{fig:sos} illustrates the token $<$EOS$>$ mapped to its corresponding vector. For clarity, the traditional sine and cosine positional encoding (PE) is utilized as shown in Eq. \ref{eq:pos}, where \(d_{\text{model}}\)=512, $i$ is the index of the dimension, and $pos$ is the token position in the input sequence. Each dimension is then represented by a wave with varying frequencies and phase offsets, with values ranging between -1 and 1 to uniquely represent each position in the sequence, as illustrated in Figure \ref{fig:pos}.

\begin{equation}
\begin{aligned}
PE_(pos, 2i) &= \sin \left( \frac{pos}{10000^{\frac{2i}{d_{\text{model}}}}} \right) \\
PE_(pos, 2i+1) &= \cos \left( \frac{pos}{10000^{\frac{2i}{d_{\text{model}}}}} \right)
\label{eq:pos}
\end{aligned}
\end{equation}

Once the input embedding is combined with the positional encodings, the resulting matrix is fed into the transformer model for the initial Q and K, while V retains only the input embedding matrix. A series of linear transformations are then applied to update Q, K, and V through weight matrices specific to each attention head. As the weight matrices are continuously learned and adjusted through backpropagation, the model’s understanding of token relations within the sequence is refined. This refined understanding is represented as the attention matrix as exemplified below in Figure \ref{fig:weight}. Each row shows how much attention one token gives to every other token, with the values in each row summing to 1. Diagonal entries tend to have higher values, which intuitively makes sense as each token attends to itself to some extent.  Similarly, “Generate” is likely to attend more to “a” (0.337) given its contextual relevance in the sequence.

\begin{figure}[ht]
\centering
\includegraphics[width=0.5\textwidth]{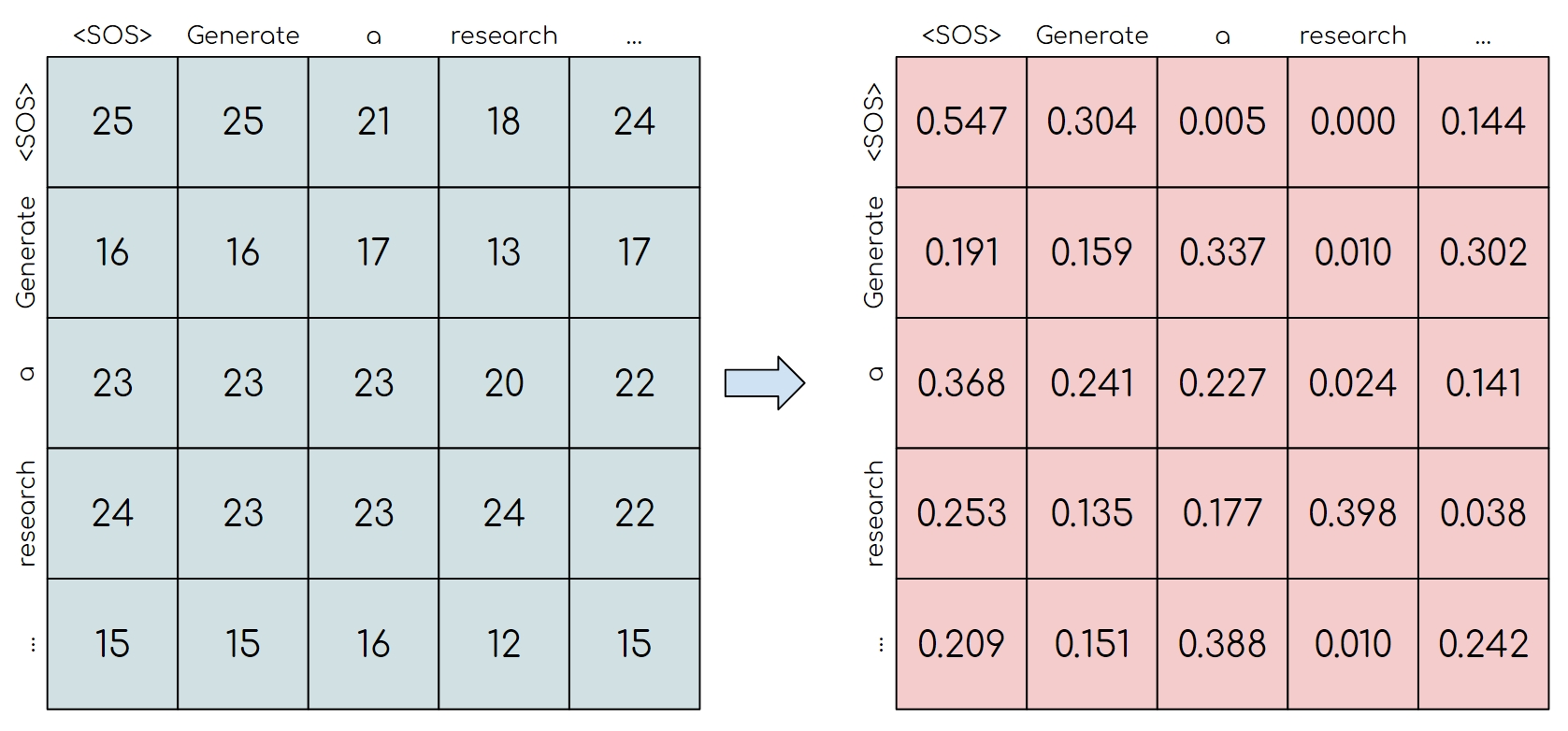}
\caption{Score and attention weight.}
\label{fig:weight}
\end{figure}

\begin{figure}[ht]
\centering
\includegraphics[width=0.15\textwidth]{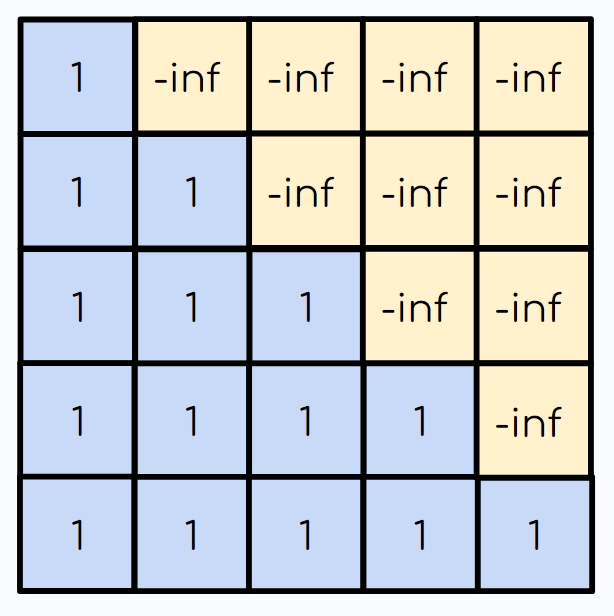}
\caption{Sample mask layer.}
\label{fig:mask}
\end{figure}

\begin{figure}[ht]
\centering
\includegraphics[width=0.5\textwidth]{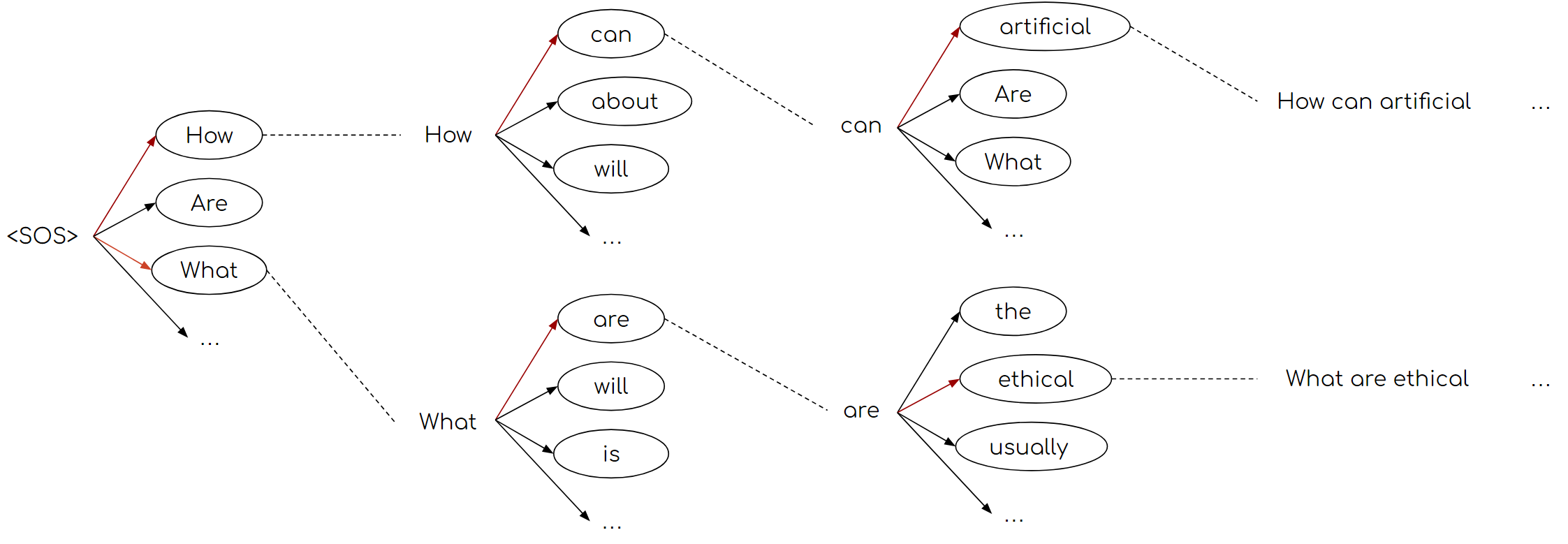}
\caption{Beam search.}
\label{fig: beam}
\end{figure}

Compared to the encoder, the decoder differs in two main aspects. It has two types of input: the encoder's output and the previously generated tokens, with the latter initialized using a special start token, $<$SOS$>$. Second, two levels of multi-head attention are applied in the decoder: masked self-attention to process the previously generated tokens, and cross-attention to focus on the encoder's output. The mask is very important, due to the sequential prediction mechanism, as the decoder cannot have a token to attend to future ones that have not yet been predicted. Therefore, future positions are masked using negative infinity as shown in Figure \ref{fig:mask}. Consequently, the query (Q) from the masked self-attention, together with the keys (K) and values (V) from the encoder’s output, undergo cross-attention to assign probabilities to future tokens in the sequence. With these probabilities, the model predicts the next token while maintaining focus on relevant portions of the input sequence. To avoid being suboptimal, beam search is employed. Instead of greedily selecting the token with the highest probability at a time, beam search keeps tracking the top k most likely sequences, where k is the width of the beam, at each decoding step. Doing so allows the model to better explore alternative output sequences and choose the one with the highest cumulative probability. In our case, the decoder might find "How" has the highest probability after $<$SOS$>$. Rather than choosing "How" immediately, the system considers the top k tokens (e.g., “Are”, “What”, etc.), and generates possible continuations for each, as shown in Figure \ref{fig: beam}. The process repeats until the decoder generates the special end token $<$EOS$>$. The sequence with the cumulative probability among all k options will be the final output sequence. 

The deep learning phase marks the official culmination of AIGC technology. The vast potential offered by neural networks has inspired researchers to achieve remarkable advancements. However, the amount of training required by neural networks and their associated high cost both in time and resources, present a significant challenge for research to further progress. The work on fine-tuning and reusing pre-trained models through transfer learning is emerging.

\section{Transfer Learning and Pre-trained Models}

As AI has evolved to date, a persistent challenge is that model training often relies heavily on large, meticulously labeled datasets and time-consuming feature engineering. These requirements make traditional deep learning methods costly and complex. To address this issue, transfer learning (TL) is designed to reuse knowledge from existing models to solve new problems. By transferring the “experience” of a pre-trained model on a large dataset to a new, relevant task, transfer learning greatly reduces the time, data, and resources required to train a model from scratch.

The basic idea of TL may seem straightforward, but understanding what and how to transfer is essential for meaningful progress. To that end, researchers have categorized TLs into four key types based on “what is delivered”. In instance-based TL, specific data points from a source domain are reused when they closely match those in a corresponding target domain. Feature-based TL transforms the features learned in the source domain so they can be applied in the target domain, even if the tasks differ, while relationship-based TL capitalizes on data relationships within the source domain to enhance performance in the target domain, particularly when shared patterns exist. Finally, model-based TL involves fine-tuning a pre-trained model from the source domain and adjusting its parameters to suit the new task’s specific needs.

It is worth noting that in addition to being categorized based on the type of knowledge transferred, TL can also be categorized from other perspectives. For instance, based on the nature of tasks, TL can be classified into isomorphic transfer\cite{5288526}, where source and target tasks are similar, and heterogeneous transfer, where they are different. Similarly, TL can be divided into supervised \cite{DBLP:journals/corr/abs-2101-05913}, semi-supervised  \cite{zhou2024continual}, and unsupervised \cite{liu-lee-2021-unsupervised} based on the level of supervision in the source and target domains. For clarity, this paper categorizes all TLs based on the type of transferred knowledge and provides details on their differences and commonalities in Subsection VII.A.

\begin{figure}[ht]
\centering
\includegraphics[width=0.5\textwidth]{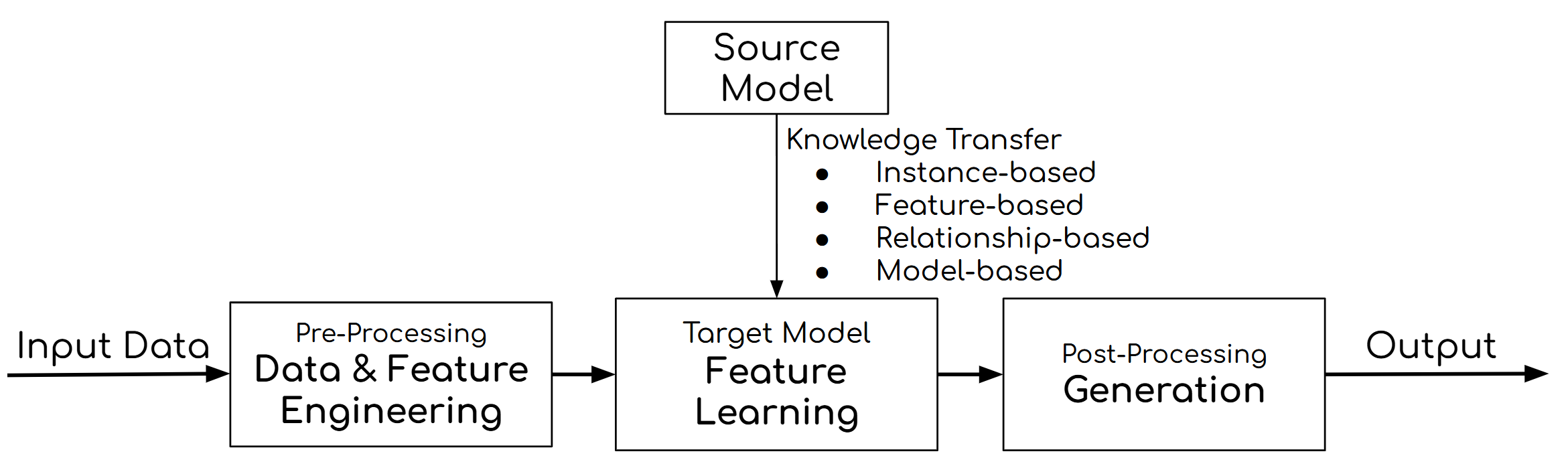}
\caption{Transfer learning prototype.}
\label{fig:fig0}
\end{figure}

After deciding what to transfer, the next focus is on "how to transfer" knowledge effectively across tasks. As shown in Figure \ref{fig:fig0}, many of the steps in TL echo those in deep learning, especially in feature pre- and post-processing. However, a key distinction in TL is the step of choosing a suitable source model. This model needs to provide useful feature representations and knowledge to support the learning of new tasks. Once the source model is chosen, its transferred knowledge is leveraged and fine-tuned to adapt to the specific requirements of the target tasks, resulting in a target model tailored to the new application. This adaptability makes transfer learning especially valuable in fields like visual and textual content generation, where creating large labeled datasets is often impractical.

\subsection{Transfer Learning Fundamentals}

In this subsection, we provide a little more detailed explanation of the four types of TL according to the type of knowledge transferred. They are instance-based TL, feature-based TL, relationship-based TL, and model-based TL. 

\subsubsection{Instance-Based TL}

Instance-based TL focuses on selectively reusing relevant samples from the source domain through different weights to enhance learning in the target domain, particularly when the two domains have overlapping data instances \cite{jiang-zhai-2007-instance}. For example, when classifying rare orchids using ample labeled data for common flowers, more similar flowers like lilies are given higher weights, while less relevant ones, like roses, are given lower weights or ignored. This approach improves generalization to the target task by identifying and weighting the most useful source instances, especially when the target domain has limited data. 

\subsubsection{Feature-Based TL}

Feature-based TL extracts transferable representations by mapping raw data from both the source and target domains into a shared feature space \cite{bengio2013representation}. For example, a pre-trained ResNet model can be used to extract features such as edges and textures from images in the source domain, which can serve as inputs for building a new face detection model in the target domain without modifying the ResNet. This process minimizes domain differences, making it ideal for scenarios where source and target domains share underlying structures, even if tasks differ \cite{daume2009frustratingly}. 

\subsubsection{Relationship-Based TL}
Instead of transferring individual features or instances, relationship-based TL focuses on relationships and dependencies between data points. This approach is especially effective for tasks where relationships are essential and the relational structures in the source and target domains are similar. For instance, a knowledge graph of side effects associated with existing drugs can be leveraged to infer potential effects for new drugs \cite{lin2015learning}. 

\subsubsection{Model-Based TL}
Model-based TL focuses on reusing the entire model architecture and its learned parameters from source tasks for targeted tasks \cite{yosinski2014transferable}. Typically, the reusable model is pre-trained on a large, related dataset. To adapt it to a new application, the early layers are frozen to retain general features while the later layers are fine-tuned for task-specific adaptation. The number of layers retrained depends on the complexity of the target task. For example, a small dataset of manufacturing defect images can be used to fine-tune a pre-trained ResNet, allowing it to recognize defect-specific patterns \cite{he2016deep}. By fine-tuning or retraining partial layers of the model, the approach significantly improves the model's efficiency and target tasks \cite{sun2017revisiting}.

\subsection{Modern Pre-trained Models: Large Language Models (LLMs)}

\textcolor{red}{With the transfer objective determined, the next key step in transfer learning is the choice of a suitable pre-trained model. These models, trained on large datasets to capture general patterns within a source domain, enable the transfer of this knowledge to a target domain. Pre-trained models are utilized across various domains, including computer vision, speech processing, and multimodal applications. This subsection focuses specifically on Large Language Models (LLMs), given their growing importance and popularity. Built upon the Transformer architecture, LLMs feature structural variations tailored to different purposes, affecting how they process information and what tasks they excel at.}


\subsubsection{Decoder-Only Models}
\textcolor{red}{Decoder-based models process information in a strictly unidirectional (left-to-right) manner using an auto-regressive attention mechanism. This configuration enables such a model to predict each token based solely on previously generated tokens, making it highly effective for generative tasks such as open-ended text generation, dialogue, and code completion. Models, such as GPT-4\cite{openai2024gpt4}, Claude \cite{claude}, PaLM\cite{PaLM}, and LLaMA \cite{touvron2023llama}, follow this design and are well-suited for scenarios requiring fluent and coherent output.}

\subsubsection{Encoder-Only Models}
\textcolor{red}{As illustrated in Section VII.C, the encoder in the Transformer architecture is responsible for understanding and extracting features from the input using bidirectional self-attention. By considering both left and right context simultaneously when encoding each token, this type of model develops a deep understanding of the input as a whole. Encoder-only models such as BERT\cite{Bert}  excel in comprehension tasks such as sentence classification, named entity recognition, extractive question answering, and retrieval.}

\subsubsection{Encoder–Decoder Models}
\textcolor{red}{As the name suggests, the Encoder-Decoder models combine both components of the Transformer architecture. Employing bidirectional self-attention in the encoder and unidirectional attention in the decoder, the models not only comprehensively capture the meaning of an input sequence but also ensure fluent and context-aware token generation. Models such as T5 \cite{T5} and BART \cite{bart} follow this structure design and are especially well-suited for tasks that require complex input-output mappings, such as machine translation, summarization, and text-to-text generation.}

\subsection{Example}
\textcolor{red}{The emergence of LLM provides
a solid foundation for advancing TL techniques. In this subsection, we use Llama 3 to exemplify the TL process in responding to the same prompt: "Generate a research question with the keywords: artificial intelligence, healthcare, and ethical implications.".}

To achieve 'Transfer', Llama 3 has two important transfer steps: pre-training and \textcolor{red}{fine-tuning}. Llama 3 is pre-trained on a massive dataset. The process allows it to grasp grammar, semantics, and even some world knowledge. This aligns with Model-Based TL, where the transformer architecture itself facilitates knowledge transfer to various downstream tasks. During pre-training, Llama 3 also extracts transferable feature representations, such as token embeddings and positional encodings (RoPE\cite{rope}). These features capture universal linguistic patterns, laying the groundwork for subsequent fine-tuning.


While pre-training provides a foundation, the fine-tuning step, on the other hand, achieves a specialization in the problematic research area. Llama 3 usually hones in on datasets related to AI and ethics. The fine-tuning process refines its vast knowledge “library” by deepening its understanding of these areas, and also extends to the unfamiliar healthcare field. After selecting a proper framework like LoRA or QLoRA\cite{qlora}, the source model is adapted to the new target field with our fed healthcare-related dataset. This process also grasps the nuances, such as the specific applications of AI in healthcare (e.g., diagnosis and treatment), and the ethical challenges they pose (e.g., bias and privacy issues). In addition, fine-tuning strengthens connections within the knowledge base, also known as relationship-based transfer, helping Llama 3 identify and prioritize relevant information. Compared to previous stages' examples, Llama 3 learns to recognize the intersection of AI, healthcare, and ethics, allowing it to generate research questions that focus precisely on these critical and interdisciplinary topics.


When prompted, Llama 3 generates a response using a combination of pre-trained and fine-tuned knowledge. First, it identifies key concepts in the request, such as “artificial intelligence,” “healthcare,” and “ethical implications,” and considers them as core elements. It then formulates a question based on its understanding of the structure of the research question and the relationship between these keywords. The specific steps in the generation process are very similar to deep learning's decoder, but at the same time, the optimization algorithms such as RoPE \cite{rope} and GQA \cite{gqa} are used to ensure the efficiency of the operation of the LLM. This process makes it possible to ask a relevant and insightful question, such as “What are the ethical implications of using AI for personalized healthcare treatment?".

In summary, TL provides a framework for knowledge adaptation and application. This enriches the deep learning ecosystem by reusing existing systems. It provides a way to overcome the limitations of deep learning by transferring and adapting pre-trained and learned models to cope with new requirements.

\section{AIGC limitations, challenges, and future trends}
AIGC technology has been under development for quite some time, and it is now accessible and usable by a wide range of users, demonstrating undeniable potential. However, AIGC presents several limitations and challenges that pose barriers to its continued advancement. In this section, we examine these issues and explore future prospects for overcoming them. Specifically, we categorize the challenges into two primary directions: data-related issues and model structure limitations, discussed in Subsections A and B, respectively.


\subsection{Data-related limitations and challenges} 

As elaborated earlier, data is the key for AIGC regardless of which types of methods are used for content creation. The greater the volume and dimensions of data, the more knowledge methods can gather, and the better the content to be created. Dependency on data is also the source of many limitations and challenges.

First, the quality of AIGC is constrained by how much data the methods have “seen” and understand. The availability and accessibility of data remain the barriers that current AIGC struggles to overcome, while research advancements in areas, such as transfer learning outlined in Section VII, few-shot\cite{alayrac2022flamingo}, and zero-shot learning \cite{zeroshot}, strive to reduce this reliance. \textcolor{red}{These methods are specifically designed to help generative models sustain high performance with minimal new data. In this context, few-shot learning allows models to generalize from a small number of labeled examples, while zero-shot learning supports task execution without any task-specific training data. In AIGC, these techniques enable applications such as summarization, question answering, and image synthesis, especially in domains with scarce or no annotated data.}

Inevitably, biases, incorrect or even inappropriate content \cite{bias} might be present in the available training data, causing AIGC systems to perpetuate misinformation or unfair results. Therefore, data augmentation, such as data filtering \cite{datafilter}, resampling \cite{smote}, adversarial training\cite{dataadtrain}, synthetic data generation \cite{gan}, and data validation \cite{crossvalid}, is essential for diverse and balanced training datasets.
\textcolor{red}{Additionally, emerging strategies for improving data quality and aligning LLMs with human preferences, such as Direct Preference Optimization (DPO) and OpenAI’s Reinforcement Learning from Human Feedback (RLHF), merit further investigation.}
Finally, establishing ethical guidelines and legal frameworks \cite{Ethical} and conducting regular bias audits, although out of the scope of this paper, are also crucial to mitigating risks and ensuring fairness in AIGC.

Training data only presents past information, which can cause generated content to appear right but is in fact wrong, leading to hallucinations. That is because AIGC models can’t access up-to-date knowledge \cite{AIGCData} during training. Recent development in Retrieval Augmented Generation (RAG) presents a promising solution \cite{rag}. By combining the strengths of retrieval-based and generation-based approaches, models’ generative ability is enhanced, extremely critical for applications, such as in fields like law and medicine, where real-time factual accuracy is essential. 

\textcolor{red}{Overall, data is still the most fundamental bottleneck in the development of AIGC, driving its future progress along two complementary directions. On the one hand, there is a growing effort to develop more data-efficient models that can achieve strong generative performance with fewer resources. Approaches such as transfer learning, few-shot, and zero-shot learning represent key strategies in this direction. 
On the other hand, expanding and improving the quality of training data is equally critical. Beyond real-world datasets, synthetic data, as a renewable “fuel” for AI, is gaining increasing attention for training and fine-tuning generative models. However, this trend also reveals new challenges: Can such data truly sustain the AIGC ecosystem over the long term? Without proper safeguards, synthetic data risks amplifying existing biases \cite{amplifyingbias}, reinforcing hallucinations \cite{hallucinations}, and even leading to “Habsburg AI” effects, such as performance degradation and systemic collapse \cite{datadegradation}. To address these challenges, future research should focus on better integration of synthetic and real-world data with stronger human oversight, both during data curation and model alignment. This includes enhanced tools for data auditing and traceability \cite{ paullada2021data}, as well as human-in-the-loop methods such as DPO that align model outputs with human intent. These efforts are ultimately essential to building high-quality, low-bias, and highly controllable datasets.}

\subsection{Model-structure-related limitations and challenges}

As AIGC evolves from early rule-based systems to deep learning, the size and sophistication of models have been increasing dramatically to improve content generation. The growth in model structures demands more processing power, memory, and storage to handle vast amounts of data necessary for both training and real-time operations. Such a large capital investment in time and power limits the widespread adoption of AIGC. Although there are no immediate solutions, some potential long-term directions include advancements in hardware accelerators (e.g., GPUs and TPUs \cite{TPU}), memory storage technologies, and renewable energy to lower computational resource consumption and costs. Recent efforts in model optimization\cite{zhang2019selfattentiongenerativeadversarialnetworks} and decentralized training across multiple devices with local data \cite{Federated} present promising avenues for reducing computational burdens. \textcolor{red}{For example, the Mixture of Experts(MoE) architecture—a recent variation of LLMs—introduces a scalable and efficient mechanism to reduce computation by activating a few lightweight "expert" subnetworks for each input via a dynamic gating mechanism. In a similar fashion, within the context of LLMs, parameter-efficient fine-tuning methods such as LoRA allow models to be adapted to new tasks by training only a small subset of parameters to cut down memory and compute requirements.}

Deep learning is often viewed as a “black box” \cite{blackbox}. This lack of transparency makes it difficult for users to understand and trust its processes and outcomes fully. As a result, deep-learning-based AIGC faces similar trustworthiness issues, and the reduced confidence in generated content limits its broader acceptance. However, ongoing research on explainable AI\textcolor{red}{\cite{XDNN}} and trustworthy AI\textcolor{red}{\cite{Trust}} is offering promising solutions to address these challenges.

Similarly, machine learning models are vulnerable to the introduction of adversarial or malicious data, which degrades their performance over time. This vulnerability extends to AIGC systems, as they rely heavily on these machine learning models. Adversarial attacks could subtly manipulate the input data to mislead model generation or directly distort the outputs. In either case, the result is incorrect, harmful, or biased content. While there has been limited work on integrating defense techniques into AIGC, the trustworthiness and security of generated content are becoming increasingly important as AIGC is applied in sensitive areas such as news generation, education, healthcare, and automated legal drafting. Techniques like adversarial training (e.g., Fast Gradient Sign Method \cite{FGSM} and Projected Gradient Descent \cite{PGD}) and robust optimization \cite{RobustOp}, which have been extensively used in machine learning, are worth exploring in future developments of AIGC to ensure more secure and reliable content generation.

Finally, ensuring coherence and relevance in AIGC while preserving its creativity is a significant challenge and a long-term goal for its development. While there are no definitive solutions yet, we believe that human-in-the-loop AIGC is the way forward. As we progress toward a human-centric industry 5.0, improved human-machine collaboration will lead to higher-quality and more innovative generated content, making it an ideal approach to achieve AIGC’s essential objectives.

\textcolor{red}{The three main challenges of AIGC model structures—scalability, opacity, and vulnerability—continue to impact its broader adoption. Future research should shift focus from merely increasing model size and power to enhancing controllability, explainability, and robustness. Promising directions include modular architectures like MoE for efficient scaling, and parameter-adaptive methods, such as LoRA, for streamlined fine-tuning. Integrative approaches like neuro-symbolic integration \cite{NeuroSymbolic}, though still emerging in the AIGC domain, also hold potential to move generative systems beyond black-box behavior towards more explainable architectures by combining neural networks with interpretable symbolic reasoning. In other words, the creation of lightweight, energy-efficient, and general-purpose frameworks is crucial to make AIGC systems accessible to everyone.}

\subsection{Beyond technique issues}

In addition to these core technical limitations, there are ethical and policy-related concerns. AIGC systems can be exploited for malicious activities, such as scams and phishing\cite{scam}, while issues like copyright disputes\cite{copyright}, data privacy\cite{aisafety}, and academic integrity continue to emerge. The ability of AIGC to amplify biases in training data, compromise personal information, and blur the lines between originality and plagiarism poses significant challenges. These ethical and technical challenges highlight the need for strong legal frameworks, ethical guidelines, and technological safeguards to ensure responsible AI deployment and development aligned with societal values. Collaboration among industry, academia, and regulators is essential to establish principles of fairness, accountability, and transparency while remaining adaptable to technological advancements.


\textcolor{red}{In summary, AIGC raises a range of ethical and policy challenges, including misuse, privacy risks, copyright issues, biases, and academic integrity concerns. These problems highlight the urgent need for legal rules\cite{Ethical}, ethical standards\cite{bommasani2021opportunities}, and technical safeguards\cite{safeguards} that grow alongside the technology. To support responsible development, principles such as fairness, transparency\cite{floridi2023ethics}, and accountability must be built into AIGC systems from the start. Addressing these challenges in a timely and proactive way will help ensure that AIGC benefits society while reducing risks. Strong governance, aligned with technical progress, is essential for building trust and ensuring that generative AI is used safely and fairly.}

\section{Conclusion}

This comprehensive survey reviews the development of AIGC, charting its evolution from basic rule-based systems to the sophisticated deep and transfer learning models that define the current state of AIGC. By dissecting the various developments in statistical modeling, deep learning innovations, and the emerging field of transfer learning, we also shed light on key breakthroughs in expanding generative AI capabilities and applications.

In addressing the constraints and challenges of the AIGC development process, we point out the multifaceted challenges and ethical dilemmas posed by the rapid evolution of AIGC technology. From longstanding struggles with data bias and model interpretability to new threats to copyright integrity and content authenticity, we emphasize the importance of technological breakthroughs and framework upgrades and suggest that stakeholders must work together to develop guidelines to ensure responsible development and deployment of AIGC technologies.

This paper aims to provide readers with a nuanced understanding of AIGC's history, current status, and future prospects. By highlighting the technological focus at different milestones, we offer insights into the progress and challenges of AIGC. Our hope is to inspire collective efforts toward shaping a future for AIGC that is not only technically sound but also trustworthy, equitable, and human-centered. It is with this expectation that we invite further research, dialog, and innovation in the evolving field of AI-generated content.


%


\ifCLASSOPTIONcaptionsoff
  \newpage
\fi



\bibliography{aigc}
\bibliographystyle{IEEEtran}

\begin{IEEEbiography}
[{\includegraphics[width=1in,height=1.25in,clip,keepaspectratio]{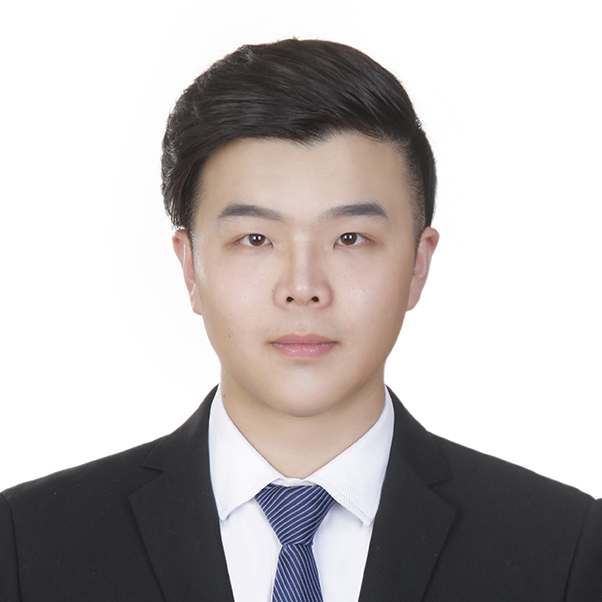}}] 
{Chengzhang Zhu} (Member, IEEE)received the B.S. degree from Nanjing University of Science and Technology, P.R. China, in 2015, and the M.S. degree from Monmouth University, USA, in 2021. He is currently pursuing the Ph.D. degree in Electrical and Computer Engineering at Rowan University, USA. His research interests include machine learning, artificial intelligence, and educational games. He has also researched generative agents and explored the development of educational games to enhance student comprehension in science, technology, and engineering fields.
\end{IEEEbiography}

\begin{IEEEbiography}
[{\includegraphics[width=1in,height=1.25in,clip,keepaspectratio]{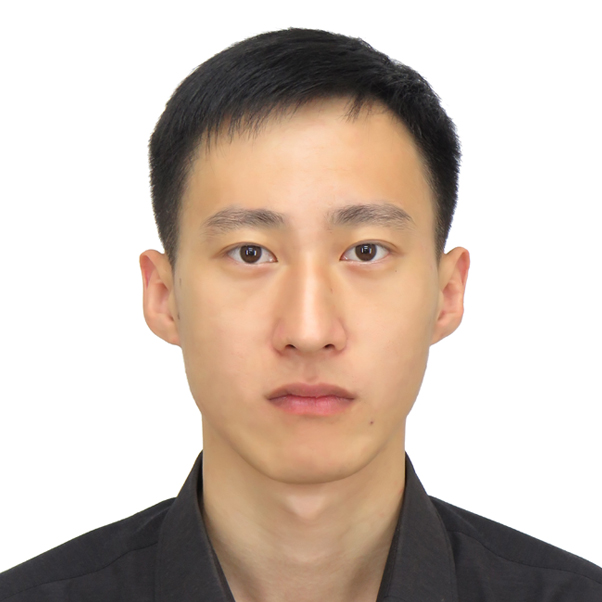}}] 
{LuoBin Cui} (Member, IEEE)received the B.S. degree from Jiangsu University of Technology, P.R. China, in 2020, and the M.S. degree from Monmouth University, USA, in 2022. He is currently pursuing the Ph.D. degree in Electrical and Computer Engineering at Rowan University, USA. His research interests include machine learning, and artificial intelligence, with a particular focus on Graph Neural Networks and their applications. He has also conducted research on fatigue detection using physiological data and explored the development of educational games to enhance student comprehension in science, technology, and engineering fields.
\end{IEEEbiography}

\begin{IEEEbiography}
[{\includegraphics[width=1in,height=1.25in,clip,keepaspectratio]{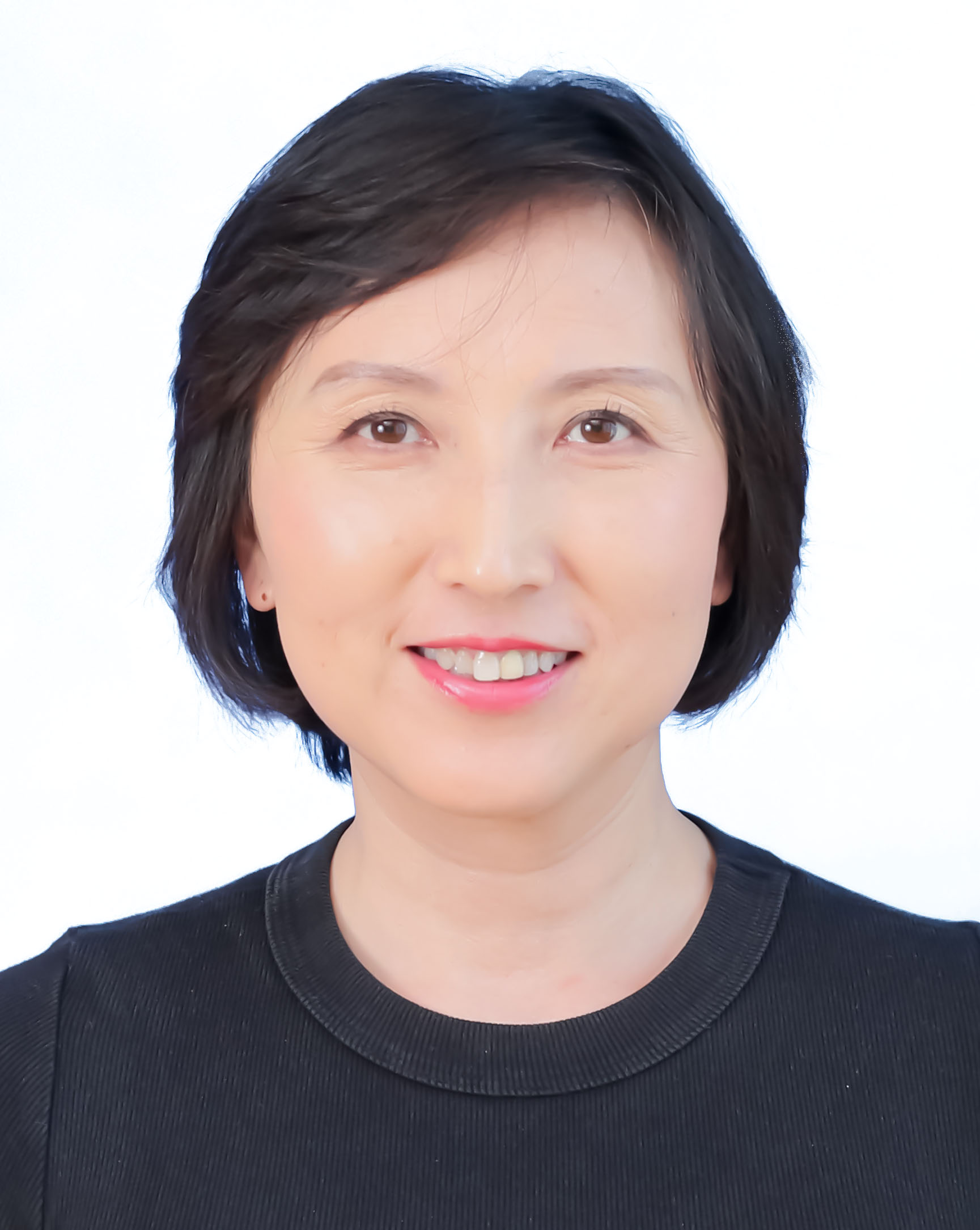}}] 
{Ying Tang} (Senior Member, IEEE) received the B.S. and M.S. degrees from the Northeastern University, P. R. China, in 1996 and 1998, respectively, and Ph. D degree from New Jersey Institute of Technology in 2001. She is currently Full Professor and the Undergraduate Program Chair of Electrical and Computer Engineering at Rowan University, Glassboro, New Jersey. Her current research interest lies in the area of cyber-physical social systems, extended reality, adaptive and personalized systems, modeling and adaptive control for computer-integrated systems, and sustainable production automation. Her work has been continuously supported by NSF, EPA, US Army, FAA, DOT, private foundations, and industry. She has three USA patents, and over 250 peer-reviewed publications, including 88 journal articles, 2 edited books, and 6 book/ encyclopedia chapters. Dr. Tang is presently Associate Editor of IEEE Transactions on Systems, Man, and Cybernetics: Systems,  IEEE Transactions on Intelligent Vehicles, IEEE Transactions on Computational Social Systems and Springer’s Discover Artificial Intelligence. She is the Founding Chair of Technical Committee on Intelligent Solutions to Human-aware Sustainability for IEEE Systems, Man, \& Cybernetic, and the Founding Chair of Technical Committee on Sustainable Production Automation for IEEE Robotic and Automation.
\end{IEEEbiography}

\begin{IEEEbiography}[{\includegraphics
[width=1in , height=1.25in,clip,
keepaspectratio]{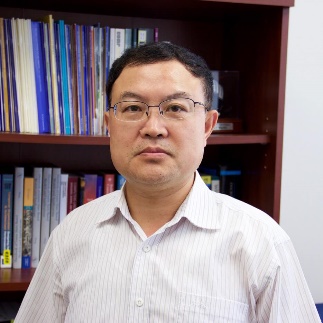}}]
{Jiacun Wang}
(SM’00) received the Ph.D. degree in computer engineering from Nanjing University of Science and Technology (NUST), China, in 1991. He is currently a Professor of software engineering at Monmouth University, West Long Branch, New Jersey, USA. From 2001 to 2004, he was a member of scientific staff with Nortel Networks in Richardson, Texas. Prior to joining Nortel, he was a research associate of the School of Computer Science, Florida International University at Miami. His research interests include software engineering, discrete event systems, formal methods, machine learning, and real-time distributed systems. He authored Timed Petri Nets: Theory and Application Kluwer, 1998), Real-time Embedded Systems (Wiley, 2018) and Formal Methods in Computer Science (CRC, 2019), edited Handbook of Finite Stat Based Models and Applications (CRC, 2012), and published over 230 research papers in journals and conferences. Dr. Wang was an Associate Editor of IEEE Transactions on Systems, Man and Cybernetics, Part C, and is currently Associate Editor of IEEE/CAA Journal of Automatica Sinica and IEEE Transactions on Systems, Man, Cybernetics: Systems. He has served as general chair, program chair, and special sessions chair or program committee member for many international conferences. He is a BoG member of the IEEE SMC society.
\end{IEEEbiography}

\vfill

\end{document}